\documentclass[sn-standardnature,iicol,pdflatex]{my-sn-jnl}
\usepackage[T1]{fontenc}
\usepackage{times}
\usepackage{xcolor}
\usepackage{amsmath,amssymb,amsfonts,bm,bbm}

\usepackage{graphicx}
\usepackage{multirow, booktabs}

\graphicspath{{./Figures/}}

\newcommand{\wh}[1]{\widehat{#1}}

\def\mkakko#1{\left(#1\right)}
\def\ckakko#1{\left\{#1\right\}}

\def\x{\mathbf{x}}
\def\y{\mathbf{y}}
\def\c{\mathbf{c}}
\def\1{\mathbbm{1}}
\def\RR{\mathbb{R}}
\def\EE{\mathbb{E}}
\def\rr{\mathbf{r}}
\def\NN{\mathbb{N}}
\def\GG{\mathbf{G}}
\def\HV{\text{HV}}
\def\PF{\text{PF}}
\def\pf{\mathfrak{pf}}
\def\gg{\mathbf{g}}
\def\T{\textsf{T}}
\renewcommand{\ss}{\mathbf{s}}
\def\aa{\mathbf{a}}

\def\NAME#1{LC-MOPG}
\def\NAMEV#1{LC-MOPG-V}

\begin{document}

\title[RL]{Latent-Conditioned Policy Gradient for Multi-Objective Deep Reinforcement Learning}

\author*{\fnm{Takuya} \sur{Kanazawa}}\email{takuya.kanazawa.cz@hitachi.com}

\author{\fnm{Chetan} \sur{Gupta}}

\affil{\orgdiv{Industrial AI Lab}, \orgname{Hitachi America, Ltd.~R\&D}, \orgaddress{\city{Santa Clara}, \state{CA} \postcode{95054}, \country{USA}}}

\abstract{%
Sequential decision making in the real world often requires finding a good balance of conflicting objectives. In general, there exist a plethora of Pareto-optimal policies that embody different patterns of compromises between objectives, and it is technically challenging to obtain them exhaustively using deep neural networks. In this work, we propose a novel multi-objective reinforcement learning (MORL) algorithm that trains a single neural network via policy gradient to approximately obtain the entire Pareto set in a single run of training, without relying on linear scalarization of objectives. The proposed method works in both continuous and discrete action spaces with no design change of the policy network. Numerical experiments in benchmark environments demonstrate the practicality and efficacy of our approach in comparison to standard MORL baselines.}


\maketitle

\section{Introduction}

\begin{figure*}[t]
	\centering
	\includegraphics[width=.8\textwidth]{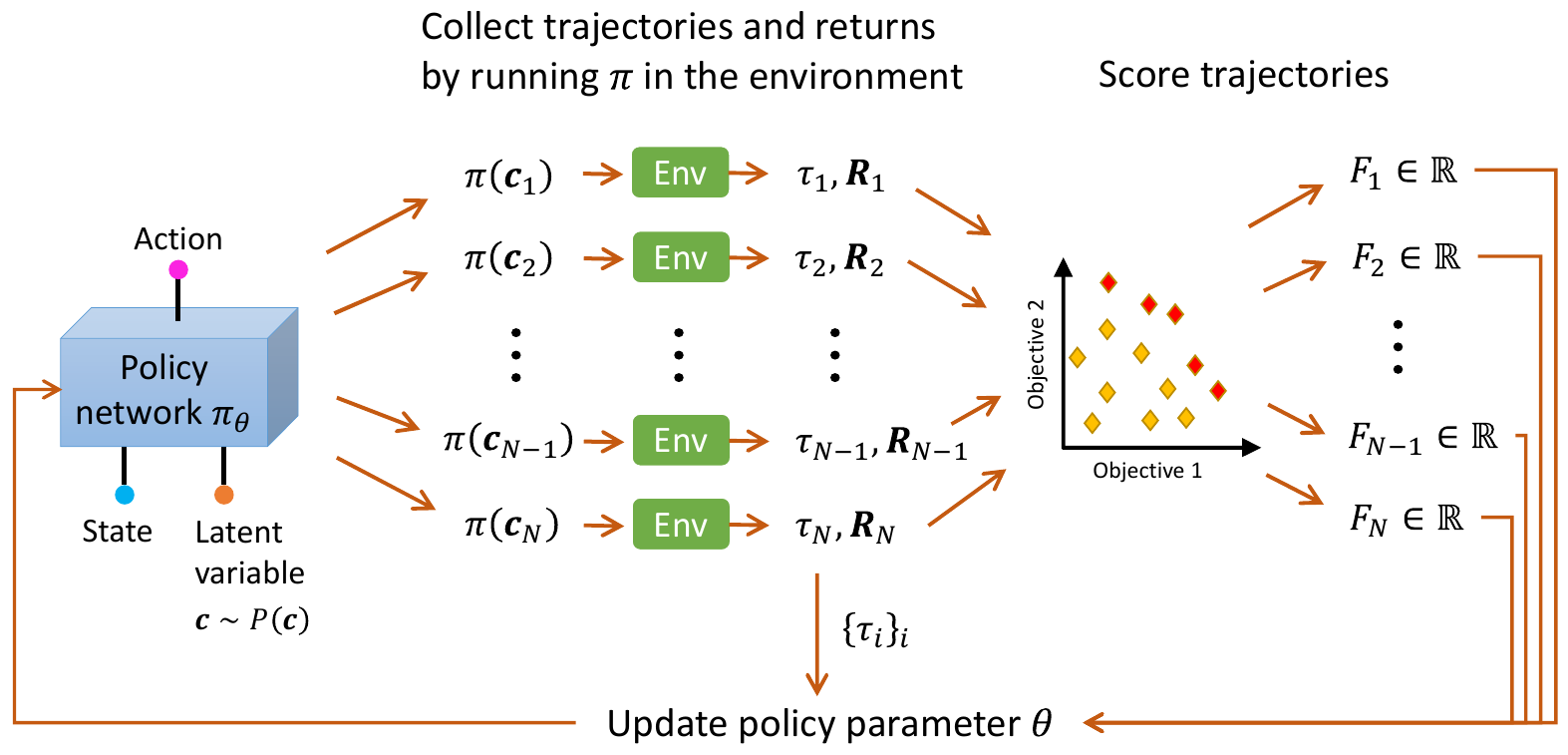}
	\vspace*{3mm}
	\caption{\label{fg_ponchie}Illustration of the proposed algorithm (\NAME{}) for a bi-objective problem. There is an alternative that uses value networks in addition to the policy network (\NAMEV{}).}
\end{figure*}

In recent years Reinforcement Learning (RL) has grown into a major part of the machine learning research. RL offers a powerful and principled framework to solve sequential decision making problems under uncertainty \cite{DBLP:books/lib/SuttonB98}. In RL, an agent takes actions sequentially in an environment and receives rewards. The goal of RL is to find an optimal policy that maximizes cumulative rewards over multiple time steps. Rapid progress in the field of deep neural networks (NN) has enabled deep RL to solve a number of complex nonlinear control tasks in the field of games, robotics, autonomous driving, traffic signal control, recommender systems, finance, etc.~at a level comparable to or even beyond human experts \cite{DBLP:journals/nature/MnihKSRVBGRFOPB15,DBLP:journals/nature/SilverHMGSDSAPL16,DBLP:journals/corr/abs-1810-06339}.

While standard RL assumes scalar rewards, RL can be extended to incorporate vector-valued rewards, which is known as Multi-Objective Reinforcement Learning (MORL) \cite{DBLP:journals/jair/RoijersVWD13,DBLP:journals/tsmc/LiuXH15,DBLP:journals/aamas/HayesRBKMRVZDHH22}. Such an extension is motivated by the fact that many real-world control problems require balancing multiple conflicting objectives; an example is a control of a walking robot where one may wish to maximize the walking speed while minimizing the electricity consumption. The existence of a trade-off between objectives naturally leads to an \emph{ensemble} of best solutions (known as \emph{Pareto-optimal policies}), each of which is characterized by an inequivalent compromise of objectives. The goal of MORL is to obtain a set of policies that approximates such an ensemble as closely as possible.  

In MORL problems, the set of true Pareto-optimal policies often consists of many (even \emph{infinite}) diverse and disparate solutions, and poses computational as well as conceptual challenges that hamper a na\"ive application of conventional RL techniques. One of the popular approaches to tackling this difficulty is to convert the original MORL problem into a series of single-objective RL problems and solve them by training an ensemble of agents \cite{DBLP:journals/corr/MossalamARW16}. This method is simple but suffers from a high computational cost. Another popular method is to train a single agent that receives a preference (linear weight vector) over objectives as additional input to the value function \cite{DBLP:conf/ijcnn/CastellettiPR12,Castelletti2013WRR}. This approach is computationally efficient but linear scalarization hampers finding the concave part of the Pareto frontier (PF) \cite{Das1997,10.1007/978-3-540-89378-3_37}.

In this work, we propose a novel model-free on-policy MORL algorithm that obtains infinitely many inequivalent policies by training just a single policy network. In this approach, as illustrated in Fig.~\ref{fg_ponchie}, a stochastic policy receives a random latent variable sampled from a fixed external probability distribution as additional input. We train this latent-conditioned NN in a policy-gradient manner, so that the quality of the set of returns generated by policies conditioned on different random latent variables gets improved successively. We introduce a novel exploration bonus that helps to substantially enhance diversity of the policy ensemble. The proposed method, coined as \underline{L}atent-\underline{C}onditioned \underline{M}ulti-\underline{O}bjective \underline{P}olicy \underline{G}radient (\NAME{}), is applicable to both continuous and discrete action spaces, and can in principle discover the whole PF without any convexity assumptions, as \NAME{} does not rely on linear scalarization of objectives. We confirm the effectiveness of \NAME{} in various benchmark tests through comparison with exact PF and standard baselines for MORL. 

The rest of this paper is organized as follows. In Sec.~\ref{sc:2334r} related work in MORL is reviewed. In Sec.~\ref{sc:hdfnv0vpo} the preliminaries are summarized and the problem to be solved is formally stated. In Sec.~\ref{sc:pqpewepq} our new method is introduced and explained in great detail. In Sec.~\ref{sc:gdf8g7wzz} the features of each environment used for benchmarking are described. In Sec.~\ref{sc:453rewfd} the results of numerical experiments are discussed and the proposed method is compared with baselines. In Sec.~\ref{sc:452sdfgz} we conclude.

\section{Related work\label{sc:2334r}}
\subsection{Scalarization-based MORL}

There are a number of MORL methods that employ some parametric scalarization function to combine multiple objectives into a single objective, and then optimize it while varying the parameters of the scalarization function so as to cover the whole PF as widely as possible. For example, Van Moffaert et al.~\cite{Moffaert2013sc} proposed the use of the Chebyshev scalarization function. This method has the advantage that it can access the entire PF without prior assumptions on its shape, but the nonlinearity of the Chebyshev function makes it impossible to derive the Bellman optimality equation. Linear scalarization (i.e., taking a weighted sum of objectives as the target of optimization) is an alternative approach, which is more popular due mainly to the fact that the Bellman equation for a weighted sum of multi-objective returns can be derived straightforwardly and readily integrated with common methodologies developed for single-objective RL. 
For example, Castelletti et al.~\cite{DBLP:conf/ijcnn/CastellettiPR12,Castelletti2013WRR} adapted fitted Q iteration to the multi-objective domain by postulating a generalized scalar Q function which receives not only a state and action pair but also a linear weight (preference) vector as input. 
While Castelletti et al.~\cite{DBLP:conf/ijcnn/CastellettiPR12,Castelletti2013WRR} used decision tree-based models to parametrize the Q function, most of more recent studies (reviewed below) use deep NN as a universal function approximator. 
Mossalam et al.~\cite{DBLP:journals/corr/MossalamARW16} introduced a NN architecture with vector output that parameterizes the multi-objective return. Training of multiple Q networks associated with different weights is done sequentially, reusing the parameters of the already trained NN to accelerate training of a new NN.
Abels et al.~\cite{DBLP:conf/icml/AbelsRLNS19} investigated the MORL setup in which a linear weight in the scalarized objective changes dynamically during an episode. Abels et al.~combined the methods of Castelletti et al.~\cite{DBLP:conf/ijcnn/CastellettiPR12,Castelletti2013WRR} and Mossalam et al.~\cite{DBLP:journals/corr/MossalamARW16} by proposing to train a single vector-valued NN that can learn the multi-objective Q function for the entire weight space efficiently. 
Yang et al.~\cite{DBLP:conf/nips/YangSN19} further improved this by introducing a new Bellman update scheme, called \emph{envelope Q-learning}, which optimizes the Q value not only over actions but also over weights, leading to more optimistic Q-updates. 
Recently, Basaklar et al.~\cite{DBLP:journals/corr/abs-2208-07914} introduced a Q-learning scheme that contains a cosine similarity term between projected weight vectors and Q values. This method requires solving a series of single-objective RL problems for fixed key weights separately, before training the agent for general weights. 

While the methods outlined above are value-based, there are also policy-based approaches for MORL that combine policy gradient \cite{Sutton1999} with linear scalarization. For example, Chen et al.~\cite{Chen2019} proposed to train a meta-policy that can approximately solve an ensemble of tasks characterized by different preferences of objectives. The meta-policy is trained with policy gradient, in which the advantage function is computed as a weighted sum of advantage functions for individual objectives. The learnt meta-policy is fine-tuned for each task separately. 
Xu et al.~\cite{DBLP:conf/icml/XuTMRSM20} proposed an evolutionary algorithm in which a population of policies are trained in parallel using various weighted sums of objectives. At each generation, the combination of a policy and a weight that best improves the PF is selected via a greedy optimization algorithm. Each policy is then trained via multi-objective policy gradient, analogously to Chen et al.~\cite{Chen2019}. 
Siddique et al.~\cite{DBLP:journals/corr/abs-2008-07773} also introduced a similar policy gradient approach. 

These policy-based approaches \cite{Chen2019,DBLP:conf/icml/XuTMRSM20,DBLP:journals/corr/abs-2008-07773} require training an ensemble of policy networks to encompass the PF, so they are prone to relatively high computational costs and implementational complexity. In this regard, we believe that our work is the first policy-based approach for MORL that efficiently learns multiple policies by training a single NN. 

\subsection{MORL without explicit scalarization}

Miscellaneous MORL approaches that do not rest on scalarization techniques have been developed. Van Moffaert and Now{\'{e}} \cite{DBLP:journals/jmlr/MoffaertN14} developed \emph{Pareto Q-learning}, which searches for Pareto-optimal policies using set-based temporal-different learning. However its application so far has been limited to small-scale problems \cite{Reymond2019}. Lexicographic RL \cite{DBLP:conf/icml/GaborKS98,DBLP:conf/ijcai/WrayZ15,DBLP:conf/ijcai/SkalseHGA22} uses lexicographic ordering of objectives, which is useful when the priority over objectives is clearly known. Reymond et al.~\cite{DBLP:conf/atal/ReymondBN22} extended the framework of reward-conditioned policies \cite{DBLP:journals/corr/abs-1912-02875,DBLP:journals/corr/abs-1912-02877,DBLP:journals/corr/abs-1912-13465} to MORL, in which the PF is learnt in a supervised learning style. Abdolmaleki et al.~\cite{DBLP:conf/icml/AbdolmalekiHNS20} proposed to learn an action distribution for each objective based on a scale-invariant encoding of preferences across objectives, extending the prior work on maximum a posteriori policy optimization \cite{DBLP:conf/iclr/AbdolmalekiSTMH18} to the multi-objective domain. 
Parisi et al.~\cite{DBLP:conf/ijcnn/ParisiPSBR14} developed two policy gradient methods, called \emph{Radial Algorithm} (RA) and \emph{Pareto-Following Algorithm} (PFA). RA assigns a set of weights to policies and optimizes each policy separately via policy gradient. PFA first performs a single-objective optimization and then updates the resulting policy incrementally to sweep the PF. PFA is not applicable to problems with more than two objectives. 

Pirotta et al.~\cite{DBLP:conf/aaai/PirottaPR15,DBLP:journals/jair/ParisiPR16} proposed a manifold-based policy search algorithm in which a policy is sampled from a continuous manifold in the policy parameter space. This manifold is specified by a parametric map from a predefined base space. The parameters of this map are updated in a gradient-descent fashion, based on a differentiable loss (or indicator) function that measures the quality of the PF. To define the loss, we have to specify several hyperparameters including coordinates of the utopia and anti-utopia points in the objective space, which requires some domain knowledge of the problem. The resulting performance was observed to be highly sensitive to the choice of these hyperparameters and calls for careful manual tuning, which limits the utility of this approach. 

Parisi et al.~proposed a generalization of natural evolution strategies (NES)  \cite{DBLP:journals/jmlr/WierstraSGSPS14} to MORL called MO-NES \cite{DBLP:journals/ijon/ParisiPP17}. Analogously to \cite{DBLP:conf/aaai/PirottaPR15,DBLP:journals/jair/ParisiPR16}, policies are sampled from a continuous parametric distribution in the policy parameter space. The parameters of this distribution are updated via NES by using the Fisher Information Matrix (FIM). MO-NES can employ a non-differentiable indicator function such as the hypervolume metric \cite{DBLP:journals/ml/VamplewDBID11}. However, a downside of MO-NES is that the number of parameters to be optimized increases rapidly with the complexity of the policy. For instance, in order to learn a distribution over policies each having 30 parameters, one has to optimize 7448 parameters in the distribution \cite{DBLP:journals/ijon/ParisiPP17}. Another downside is that, to make the computation of FIM feasible, the form of the distribution must be simple enough (e.g., Gaussian), which limits the flexibility of this approach.

Our algorithm, \NAME{}, is similar to \cite{DBLP:conf/aaai/PirottaPR15,DBLP:journals/jair/ParisiPR16,DBLP:journals/ijon/ParisiPP17} in that a continuous PF (i.e., infinitely many policies) can be learnt in a single run of training. There are, however, some important advantages for \NAME{}. First, unlike \cite{DBLP:conf/aaai/PirottaPR15,DBLP:journals/jair/ParisiPR16,DBLP:journals/ijon/ParisiPP17}, \NAME{} is a nonparametric approach, implying that we need not manually design the parametric form of the distribution over policies. Secondly, \NAME{} does not require specifying reference points (utopia and anti-utopia) in the objective space, obviating the need for domain knowledge of the desired solutions. Thirdly, \NAME{} makes it possible to train deep policies with thousands of parameters efficiently over a high-dimensional continuous state space.

\subsection{Implicit generative networks}

The concept of \emph{universal value function approximators} (UVFA) was introduced by Schaul et al.~\cite{DBLP:conf/icml/SchaulHGS15}. UVFA postulate a single NN for a value function that generalizes not only over states but also over goals. As UVFA enable efficient transfer of knowledge and skills from an already acquired goal to a new goal, UVFA serve as a powerful scheme to tackle generic multi-goal RL problems \cite{DBLP:journals/jair/ColasKSO22}. 

In a broader context, NN that receive a latent variable as additional input are generally called \emph{latent-conditioned NN} or \emph{implicit generative networks} (IGN). In the context of UVFA, the latent variable represents a goal. Unlike plain NN of the form $\y=f_\theta(\x)$ with $\x$ the input features and $\y$ the label, IGN is of the form $\y=f_\theta(\x,\c)$ where the latent variable $\c$ is typically sampled from a fixed (discrete or continuous) probability distribution such as $\mathcal{N}(\mathbf{0},\1)$. For a given $\x$, one can generate arbitrarily many distinct output $\{\y_i\}_{i=1}^N$ by feeding randomly sampled latent variables $\{\c_i\}_{i=1}^N$ to IGN, hence the name ``generative''. 
IGN has been employed for uncertainty quantification of deep learning models and has shown competitive performance \cite{DBLP:conf/nips/BouchacourtMN16,DBLP:conf/ijcci/Kanazawa022,DBLP:journals/corr/abs-2107-03743}. In RL, IGN is often used as an architecture for value networks that can capture the probabilistic distribution of expected future returns \cite{DBLP:conf/icml/DabneyOSM18}. Tessler et al.~\cite{DBLP:conf/nips/TesslerTM19} proposed a generalized policy gradient algorithm that adopts IGN to express a stochastic policy over a continuous action space in a nonparametric manner. Our work is similar to \cite{DBLP:conf/nips/TesslerTM19} in that both employ IGN for the policy network. The main difference, however, is that \cite{DBLP:conf/nips/TesslerTM19} uses IGN to represent a \emph{single stochastic policy} that samples a new latent variable $\c$ at each time step, whereas our work adopts IGN to represent \emph{a population of (deterministic) policies} each of which samples $\c$ at the start of an episode and uses it until the episode ends. 

Among the challenges faced by RL are the issues of sample complexity and brittleness against task variations. A promising approach to overcoming them is unsupervised skill discovery. Eysenbach et al.~\cite{DBLP:conf/iclr/EysenbachGIL19} showed that diverse behaviors (e..g, walking, jumping, and hopping of a bipedal robot) can be learnt without external rewards. Specifically, they used IGN for the policy and introduced an intrinsic reward that incentivises to maximize the mutual information between a probabilistic latent variable of IGN and states induced by a latent-conditioned policy. A large body of work has been done to improve this approach \cite{DBLP:journals/corr/abs-1807-10299,DBLP:conf/iclr/SharmaGLKH20,DBLP:conf/icml/CamposTXSGT20,DBLP:conf/nips/KumarKLF20,DBLP:journals/nn/OsaTS22,DBLP:conf/iclr/StrouseBWMH22}. It is noteworthy that the adoption of IGN has made training of a diverse set of policies significantly faster and more efficient than the preceding RL-based approach to skill discovery by Kume et al.~\cite{DBLP:journals/corr/abs-1710-06117}, which required training hundreds to thousands of NN independently. 

Neuroevolution is a competitive alternative to RL for diverse skill discovery \cite{DBLP:conf/gecco/ColasMHC20,DBLP:journals/corr/abs-2210-03516}. Unlike deep RL, neuroevolution applies evolutionary algorithms to update policies. Recent works \cite{DBLP:conf/gecco/NilssonC21,DBLP:conf/gecco/PierrotMCFCBCSP22} combined neuroevolution (specifically, Quality-Diversity methods) with policy-gradient updates for better sample efficiency.

\section{Problem formulation\label{sc:hdfnv0vpo}}
\subsection{Markov decision processes}

We consider sequential decision making problems that are formally described as a Markov Decision Process (MDP): $\langle S, A, P, r, \gamma, \rho_0 \rangle$ where $S$ and $A$ are the state and action spaces, $P: S\times A \times S \to [0,\infty)$ is the transition probability (density), $r: S\times A \times S \to \RR$ is the reward function, $\gamma\in [0,1]$ is the discount factor, and $\rho_0$ is the initial state distribution. 

A policy $\pi$ specifies which action is taken in a given state. A deterministic policy $\pi:S\to A$ yields action as a function of state, while a stochastic policy $\pi: S\times A\to [0,\infty)$ gives a probabilistic distribution over $A$ as a function of state. The value function $v_\pi: S\to \RR$ of a policy $\pi$ is formally defined as
\begin{align}
	v_{\pi}(s) & = \EE_\pi \left[
		\sum_{k=0}^{\infty}\gamma^k r_{t+k} \bigg\vert s_t=s
	\right] .
\end{align}
The (discounted) cumulative sum of rewards $G_t = \sum_{k=0}^{\infty}\gamma^k r_{t+k}$ is called \emph{return}. 
The goal of RL methods is to (approximately) find the optimal policy $\pi_*$ that maximizes $v_{\pi}(s)$ for all $s\in S$ \cite{DBLP:books/lib/SuttonB98}. 

\subsection{Policy gradient}

Policy-based methods iteratively update the parameters $\theta$ of a policy $\pi_\theta$ to improve the expected return. The classical REINFORCE algorithm \cite{Williams1992} for episodic environments runs a stochastic policy, samples trajectories $\{(s_t,a_t)\}_t$, and optimizes the parameters as
\begin{align}
	\theta \leftarrow \theta + \eta \sum_t G_t \nabla_\theta 
	\log \pi_\theta (a_t \vert s_t)
\end{align}
where $\eta>0$ is the learning rate. Intuitively this update rule implies that actions that lead to higher return are reinforced more strongly than those that lead to lower return. This method, based on the Monte Carlo estimate of return, is prone to high variance. To remedy this, we may utilize an \emph{advantage function} $A_\phi: S\times A\to \RR$, which has its own parameters and is trained concurrently with the policy via recursive Bellman updates. The policy is then optimized iteratively as
\begin{align}
	\theta \leftarrow \theta + \eta \sum_t A_\phi(s_t, a_t) \nabla_\theta 
	\log \pi_\theta (a_t \vert s_t)\,. 
\end{align}
This class of RL algorithms is called (on-policy) Actor-Critic methods \cite{DBLP:conf/nips/KondaT99}. Widely used variants are A3C \cite{DBLP:conf/icml/MnihBMGLHSK16}, TRPO \cite{DBLP:conf/icml/SchulmanLAJM15}, and PPO \cite{DBLP:journals/corr/SchulmanWDRK17}.

\subsection{Multi-objective MDP}

MDP may be generalized to multi-objective MDP (MOMDP) for multi-objective sequential decision making problems, in which the reward function $\rr: S\times A \times S\to \RR^m$ becomes vector-valued, with $m\geq 2$ the number of objectives. The discount factor $\gamma$ can also be generalized to $m$ components, but for the sake of simplicity, we will limit ourselves to a common $\gamma$ for all objectives throughout this paper. 

In MOMDP, a policy $\pi$ induces a vector of expected returns 
\begin{align}
	\GG^\pi & \equiv (G^\pi_1, G^\pi_2, \cdots, G^\pi_m)
	\\
	& = \EE\left[
	\sum_{t=0}^{\infty}\gamma^t \rr(s_t,a_t,s_{t+1})
	\bigg \vert
	a_t \sim \pi(s_t)
	\right]\in \RR^m\,.
\end{align}
A policy $\pi$ is said to \emph{dominate} another policy $\pi'$ if $G_i^\pi \geq G_i^{\pi'}$ for all $i$ and $G_i^\pi > G_i^{\pi'}$ for at least one $i$. (With a slight abuse of terminology, we also say that $\GG^\pi$ dominates $\GG^{\pi'}$, and express this relation as $\GG^\pi \succ \GG^{\pi'}$.) If there exists no policy that dominates $\pi$, then $\pi$ is said to be \emph{Pareto-optimal}. The set of all such policies is called the \emph{Pareto set}, and the set of expected returns of all Pareto-optimal policies is called the \emph{Pareto frontier} (PF), which is a submanifold of $\RR^m$. The goal of MORL is to find a set of policies that best approximates the true Pareto set. 

In MOMDP, there is a possibility that the Pareto set is composed of stochastic policies \cite{DBLP:journals/jair/RoijersVWD13}. Nevertheless, in real-world deployment of trained RL agents, deterministic policies are often preferred to stochastic policies because the latter are intrinsically unpredictable and quality assurance becomes nontrivial. In this paper, we will use stochastic policies in the training phase to facilitate exploration, and switch to deterministic policies in the evaluation phase.  

There are several performance metrics proposed in the literature to assess the quality of an approximate PF \cite{DBLP:conf/icse/WangAYLL16,DBLP:journals/tse/LiCY22}. Some metrics require knowing the true PF, which is often difficult in real-world applications. One of the most widely used metrics that do not require knowing the true PF is the \emph{hypervolume} indicator \cite{DBLP:journals/tec/ZitzlerT99,DBLP:journals/ml/VamplewDBID11}, which is essentially the area (volume) of regions in $\RR^m$ that are dominated by a given (approximate) PF. To make the hypervolume finite and well-defined, we need to bound $\RR^m$ by introducing a reference point, often referred to as the anti-utopia point $\GG_{\rm AU}$, which is worse than the PF in all objectives. Then the hypervolume indicator $\HV$ is defined as
\begin{align}
	\HV(\text{PF}) & = \mu\mkakko{\bigcup_{\GG^*\in\text{PF}}\Omega(\GG^*)},
	\\
	\Omega(\GG^*) & = \ckakko{\gg\in \RR^m \big \vert
	\GG_{\rm AU}\prec \gg\prec \GG^*}
\end{align}
where $\mu$ is the Lebesgue measure. HV is the only unary indicator that is known to be Pareto compliant \cite{DBLP:journals/tec/ZitzlerTLFF03}.

\section{Methodology\label{sc:pqpewepq}}

In this section, we introduce a novel policy-gradient approach to MORL coined as \NAME{}. The pseudocode is presented in Algorithm~\ref{alg:main1}. 
\begin{algorithm}[tb]
    \caption{Latent-Conditioned Multi-Objective Policy Gradient (\NAME{})}
    \label{alg:main1}
    \begin{algorithmic}[1]
        \Require 
        $\pi_\theta$: policy network, 
        $\gamma\in (0,1]$: discount rate, 
        $d_{\rm lat}\in\NN$: dimension of latent variables, 
        $\beta\geq0$: bonus coefficient, 
        $N_{\rm lat}\in\NN$: number of latent variables sampled per iteration
        \State Initialize $\pi_\theta$
        \While{$\pi_\theta$ has not converged}
            \For{$i$ in $\{1,2,\cdots,N_{\rm lat}\}$}
                \State Sample $\c_i\sim \text{Uniform}
                \mkakko{[0,1]^{d_{\rm lat}}}$
                \State Obtain trajectory $\tau_i=\{(\ss_t, \aa_t)\}_t$ and 
                \Statex \qquad \quad 
                return $\GG_i$ by running policy $\pi_\theta(\c_i)$ in 
                \Statex \qquad \quad the environment
            \EndFor
            \State Compute normalized returns 
            $\big\{\wh{\GG}_i \big\}_{i=1}^{N_{\rm lat}}$ 
            \Statex \quad ~
            from $\{\GG_i \}_{i=1}^{N_{\rm lat}}$\vspace{1mm}%
            \State Compute scores $\{f_i\}_{i=1}^{N_{\rm lat}}$ 
            from $\big\{\wh{\GG}_i \big\}_{i=1}^{N_{\rm lat}}$
            \Statex
            \hfill $\triangleright$ Algorithm~\ref{alg:scoring}
            \vspace{1mm}
            \State Compute bonuses $\{b_i\}_{i=1}^{N_{\rm lat}}$ 
            from $\big\{\wh{\GG}_i \big\}_{i=1}^{N_{\rm lat}}$
            \Statex
            \hfill $\triangleright$ Algorithm~\ref{alg:bonus}
            \vspace{1mm}
            \For{$i$ in $\{1,2,\cdots,N_{\rm lat}\}$}
                \State $F_i \gets \max(f_i + \beta b_i, 0)$
            \EndFor
            \State Update $\theta$ via gradient descent of loss
            \Statex 
            \qquad $\displaystyle \mathcal{L}_{\pi}(\theta) =
            - \sum_{i=1}^{N_{\rm lat}}\bigg\{
            F_i \sum_{(\ss,\aa)\in \tau_i}
            \log \pi_{\theta}(\aa \vert \ss,\c_i) \bigg\}$
        \EndWhile
        \\
        \Return $\pi_\theta$
    \end{algorithmic}
\end{algorithm}
The main idea is quite simple: train a latent-conditioned NN as a parameter-efficient representation of a diverse collection of policies in the Pareto set of given MOMDP. The training is done on-policy, i.e., we start with a randomly initialized policy and use actions sampled from the stochastic policy for exploration. Over the course of training the policy's stochasticity decreases, signaling convergence. Since the exploration ability of this algorithm entirely hinges on the randomness of the policy in the early phase of training, we propose to initialize the NN parameters so that the resulting initial policy is (nearly) uniformly random in the action space. 

An important point to note is that the latent variable $\c$ that is fed to the policy network carries no such intuitive meaning as ``preference over objectives.'' In the proposed algorithm $\c$ is simply a mathematical steering wheel to switch policies. We will use a uniform distribution $\text{Uniform}\mkakko{[0,1]^{d_{\rm lat}}}$ as a latent space distribution $P(\c)$, in accordance with \cite{DBLP:conf/iclr/SharmaGLKH20}. The dimension $d_{\rm lat}\in \NN$ is an important hyperparameter of the algorithm. Higher $d_{\rm lat}$ implies larger capacity (or representation power) of the policy NN, but too high $d_{\rm lat}$ makes training slow. If the dimension of the PF, dim(PF), is known, then $d_{\rm lat}$ had better be at least as large as dim(PF). We recommend using $d_{\rm lat}=\text{dim}({\rm PF})+1$ or $\text{dim}({\rm PF})+2$, which are found to work well empirically. 

\begin{figure}[t]
	\centering
	\includegraphics[width=.8\columnwidth]{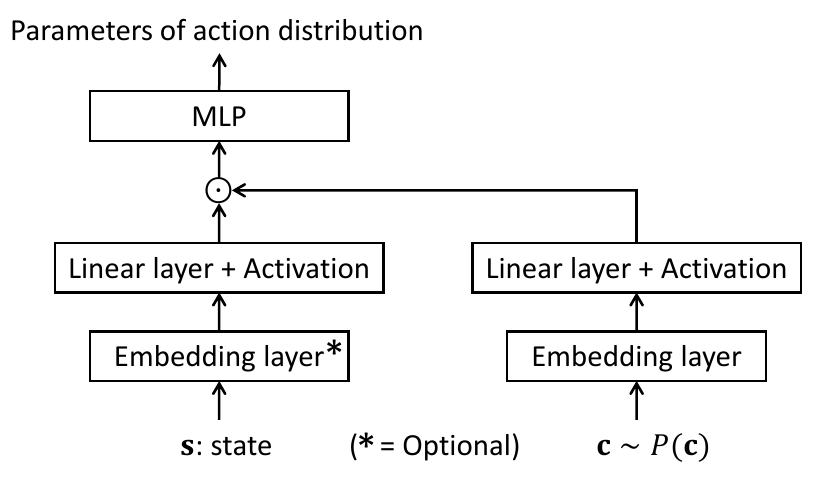}
	\caption{\label{fg_pi}Architecture of the policy network.}
\end{figure}

The policy NN used in this work has the structure shown in Fig.~\ref{fg_pi}. It provides a mapping from a state+latent space to action distribution. The outputs of policy are parameters of any parametric probability distribution that is suitable for MOMDP under consideration; e.g., the mean and variance if a Gaussian distribution is chosen. For a continuous bounded action space, we choose to use the Beta distribution $B(\alpha,\beta)$ with $\alpha,\beta>0$ for each action dimension, as recommended by Chou et al.~\cite{DBLP:conf/icml/ChouMS17}. For a discrete action space, the outputs of policy are logits of action probabilities.

The feature embedding layer for latent input, which has no trainable parameters, is basically the same as in \cite{DBLP:conf/icml/DabneyOSM18} and explicitly given as
\begin{multline}
	\big( \cos (\pi c_1), \cos (2\pi c_1), \cdots, \cos (K \pi c_1) \big) 
	\oplus \cdots 
	\\
	\cdots \oplus 
	\big( \cos (\pi c_{d_{\rm lat}}), \cos (2\pi c_{d_{\rm lat}}), \cdots, \cos (K \pi c_{d_{\rm lat}}) \big)
	\label{eq:4mbpwqf}
\end{multline}
which embeds $\c\in[0,1]^{d_{\rm lat}}$ into $[-1,1]^{K d_{\rm lat}}$. The inflation factor $K\in\NN$ is a free parameter. In principle one can also apply the same embedding to a (properly shifted and scaled) state vector; this is not mandatory and its benefit should be examined for each problem on an individual basis. Finally, the outputs of state layers and latent layers are mixed through a dot product. For image inputs, convolutional neural networks may be used as the first layer for states. 

We note that in the evaluation phase (i.e., after training is done) we make the policy deterministic. For discrete actions, this is done by simply taking the action with the highest probability. For continuous actions, the mean of the Beta distribution is selected for each dimension. 

The crux of \NAME{} is the parameter update scheme of policy. Let us recall that, in policy gradient methods for single-objective RL such as REINFORCE \cite{Williams1992}, trajectories $\{\tau_i\}_i$ can be easily classified into `good' ones and `bad' ones, according to their return values $\{G_i \in\RR\}_i$; subsequently, actions in good trajectories are reinforced while those in bad ones are weakened. In MORL, however, returns $\{\GG_i\in\RR^m\}_i$ are vector-valued and there is more than one way to assess the quality of trajectories. We now describe \NAME{}'s scheme in four steps.

\paragraph{Normalization of returns} In the first step, raw returns $\{\GG_i\in\RR^m\}_i$ obtained by running latent-conditioned policies $\pi_\theta(\c_i)$, $\c_i\sim P(\c)$, must be normalized to make the algorithm insensitive to the scale of rewards. We shall adopt one of the normalization conventions below.
\begin{itemize}
	\item Standard normalization: 
	\begin{align}
		\wh\GG = \big(\GG-\textsf{mean}(\{\GG_i \}_i)\big) 
		/ \textsf{std}(\{\GG_i \}_i)
		\label{eq:fd8gf8gfdz8}
	\end{align}
	where \textsf{mean} and \textsf{std} are 
	computed for each dimension separately. 
	The division of a vector by a vector is performed elementwise.
	\item Robust normalization:
	\begin{align}
		\wh\GG = \big(\GG-\textsf{median}(\{\GG_i \}_i)\big) 
		/ \textsf{iqr}(\{\GG_i \}_i)
		\label{eq:089df}
	\end{align}
	where \textsf{iqr} denotes the interquartile range. 
	\item Max-min normalization:
	\begin{align}
		\wh\GG = \frac{\big(\GG-\textsf{median}(\{\GG_i \}_i)\big)} 
		{\textsf{max}(\{\GG_i \}_i) - \textsf{min}(\{\GG_i \}_i)}\,.
	\end{align}
\end{itemize}
We empirically found robust normalization most useful because it is stable in the presence of outliers due to a large negative reward given to a fatal failure, such as the falling down of a robot. 

In some environments, there may exist a locally optimal solution such that most of rollout trajectories are clustered around this solution, resulting in vanishingly small denominators in \eqref{eq:fd8gf8gfdz8} and \eqref{eq:089df}. In such a case, the max-min normalization is numerically more stable.

\paragraph{Scoring of returns} The second step is score computation for each normalized return $\wh\GG_i$, as summarized in Algorithm~\ref{alg:scoring}. The main idea is to first determine the current PF ($=$ the set of undominated points) $\pf=\PF(\{\wh\GG_i \}_i)$; Next, compute the distance between $\pf$ and each $\wh\GG_i$; Finally, use the negative of this distance as a score (higher is better). This simple protocol is slightly modified in Algorithm~\ref{alg:scoring} (lines 4--6) due to the following reason.
\begin{algorithm}[tb]
    \caption{Score computation}
    \label{alg:scoring}
    \begin{algorithmic}[1]
        \Require Normalized returns 
        $\big\{\wh\GG_i \in \RR^m \big\}_{i=1}^{N_{\rm lat}}$ 
        \State Obtain the set of undominated points 
        $\pf=\PF\big(\big\{\wh\GG_i \big\}_{i}\big)$
        \For{$i\in\{1,2,\cdots,N_{\rm lat}\}$\vspace{1mm}}
            \State $\mathfrak{D}\leftarrow \ckakko{
                \underset{z\in\pf}{\min} \; 
                \| z - \wh\GG_i \|_2
            }$
            \For{$j\in\{1,2,\cdots,m\}$}
                \State 
                $\mathfrak{D} \leftarrow \mathfrak{D} \cup 
                \ckakko{
                    \underset{z\in \pf}{\max} \; z_j
                    - \big(\wh\GG_i \big)_j
                }$
            \EndFor
            \State $f_i \leftarrow - \min \mathfrak{D}$
        \EndFor
        \For{$i\in\{1,2,\cdots,N_{\rm lat}\}$}
            \State $f_i \leftarrow f_i - \textsf{avg}(\{f_i\}_i)$
            \Statex 
            \hfill $\triangleright$ \textsf{avg} $\in \{\textsf{mean}, \textsf{median}\}$
        \EndFor
        \\
        \Return Scores $\{f_i\}_{i=1}^{N_{\rm lat}}$
    \end{algorithmic}
\end{algorithm}
Let us imagine that biobjective returns are uniformly distributed inside the square $[0, 100]^2$. If we add a point $P=(100, 100)$ to this set, $P$ obviously becomes the sole PF since all points inside the square are dominated by $P$. Now, let us look at a point $Q=(0, 100)$. Since $Q$ is far away from $P$, the score of $Q$ must be very low, according to the above simple protocol. However, if $Q$ is slightly shifted upward (e.g., $Q=(0,100+10^{-4})$), $Q$ suddenly becomes a new PF and acquires the highest score. Such a discontinuous jump of scores would make the policy optimization bumpy and should be avoided. This discontinuity is removed by lines 4--6 of Algorithm~\ref{alg:scoring}.

It is a common practice in evolutionary algorithms to use robust scores such as rank-based metrics instead of raw objective values to assess the fitness of a population \cite{DBLP:journals/ec/HansenO01,DBLP:journals/jmlr/WierstraSGSPS14,DBLP:journals/ijon/ParisiPP17,DBLP:journals/corr/SalimansHCS17}. We do not employ this method because ranks of returns change discretely as we vary the policy parameters continuously and make the optimization landscape less smooth. Note also that we do not use HV-based metrics used e.g., by MO-NES \cite{DBLP:journals/ijon/ParisiPP17}, because HV is generally sensitive to the choice of a reference point.

\paragraph{Bonus computation} The third step of policy assessment in \NAME{} is \emph{bonus} computation. The bonus, denoted $b_i$, is to be added to the score $f_i$ for the sake of better exploration and higher diversity of policies. The bonus is especially useful when the scoring scheme described above alone is not enough to prevent policies from collapsing to local optima. Although our bonus bears similarity to intrinsic rewards used widely in RL \cite{Hao2023}, a notable difference is that intrinsic rewards are usually independent of external rewards $r_t$ while our bonus is directly calculated from the return distribution $\{\GG_i\}_i$ of trajectories. 

The procedure of bonus computation is outlined in Algorithm~\ref{alg:bonus}. The essential features are:
\begin{algorithm}[tb]
	\caption{Bonus computation}
    \label{alg:bonus}
	\begin{algorithmic}[1]
		\Require 
		Normalized returns $\big\{\wh\GG_i \big\}_{i=1}^{N_{\rm lat}}$, scores $\{f_i \}_{i=1}^{N_{\rm lat}}$, and parameter $k\in\NN$
		\For{$i = \{1,2,\cdots,N_{\rm lat}\}$}
			\State $\textsf{mask}_i \leftarrow 1$ if $f_i>0$ else $0$
			\State $d_i \leftarrow$ distance from $\wh\GG_i$ to its $k$th nearest 
			\Statex\qquad 
			neighbor in $\big\{\wh\GG_i \big\}_{i=1}^{N_{\rm lat}}$
			\State $b_i\leftarrow \textsf{mask}_i \ast d_i$
		\EndFor
		\\
		\Return $\{b_i \}_{i=1}^{N_{\rm lat}}$
	\end{algorithmic}
\end{algorithm}
\begin{itemize}
    \item A high bonus is given to a trajectory $i$ whose return $\wh\GG_i$ is dissimilar to other returns.
    \item Trajectories whose returns form a dense cluster in $\RR^m$ receive low bonuses. 
\end{itemize}
This distinction is quantified by the local density of returns $\big\{\wh\GG_i \big\}_{i=1}^{N_{\rm lat}}$ in $\RR^m$. We shall measure it via the $k$-nearest-neighbor distance. The index $k\in\NN$ is a hyperparameter that must be carefully tuned in conjunction with the population size $N_{\rm lat}$. It is recommended to keep $k/N_{\rm lat}$ roughly constant when $N_{\rm lat}$ is varied. Unfortunately, this na\"{i}ve definition of bonus may cause a rapid degradation of policies when there are highly inferior policies whose returns $\wh\GG_i$ are located far below those of other policies. To prevent this from happening, we multiply each bonus by a binary mask $\in\{0,1\}$ so that \emph{only trajectories in the better half of the population get bonuses} (lines 2 and 4 of Algorithm~\ref{alg:bonus}). 

\paragraph{Clipping} The fourth and last step of our policy assessment is to cut off the sum of score and bonus at zero (line 11 of Algorithm~\ref{alg:main1}). This step yields the final score of the $i$th trajectory as $F_i=\max(f_i + \beta b_i, 0)$, which is used in the subsequent policy gradient update. This clipping means that, unlike standard policy gradient methods, we do not penalize (or weaken) actions in inferior trajectories. We found this expedient to stabilize the entire learning process well.

This completes the description of the overall design and technical details of the proposed algorithm \NAME{}. 

In single-objective RL it is well known that on-policy trajectory-based methods (REINFORCE) are generally outperformed by methods with \emph{advantage} functions such as A2C \cite{DBLP:conf/icml/MnihBMGLHSK16} and PPO \cite{DBLP:journals/corr/SchulmanWDRK17}. The advantage function $A(\ss_t, \aa_t)$ enables to estimate a proper update weight for each transition (state-action pair), which therefore provides a more fine-grained policy update than a trajectory-based update. We can consider a similar generalization of \NAME{}, named \NAMEV{}. The pseudocode is presented in Algorithm~\ref{alg:main2}. 
\begin{algorithm}[h!]
    \caption{\NAME{}  
    with Generalized Value Networks (\NAMEV{})}
    \label{alg:main2}
    \begin{algorithmic}[1]
        \Require 
        $\pi_\theta$: policy network, 
        $\{Q_\phi, V_\psi \}$: generalized value networks, 
        $\gamma\in (0,1]$: discount rate, 
        $d_{\rm lat}\in\NN$: dimension of latent variables, 
        $\beta\geq0$: bonus coefficient, 
        $N_{\rm lat}\in\NN$: number of latent variables sampled per iteration, 
        $D$: rollout buffer
        \State Initialize $\pi_\theta, Q_\phi, V_\psi$
        \While{$\pi_\theta$ has not converged}
            \For{$i$ in $\{1,2,\cdots,N_{\rm lat}\}$}
                \State Sample $\c_i\sim \text{Uniform}
                \mkakko{[0,1]^{d_{\rm lat}}}$
                \State Obtain trajectory $\tau_i=\{(\ss_t, \aa_t)\}_t$ 
                and 
                \Statex \qquad 
                return $\GG_i$ by running policy $\pi_\theta(\c_i)$ in 
                \Statex \qquad 
                the environment
                \State Store $\tau_i$ in $D$ 
            \EndFor
            \State Compute normalized returns 
            $\big\{\wh{\GG}_i \big\}_{i=1}^{N_{\rm lat}}$ 
            \Statex \quad ~
            from $\{\GG_i \}_{i=1}^{N_{\rm lat}}$\vspace{1mm}%
            \State Compute scores $\{f_i\}_{i=1}^{N_{\rm lat}}$ 
            from $\big\{\wh{\GG}_i \big\}_{i=1}^{N_{\rm lat}}$
            \vspace{1mm}
            \State Compute bonuses $\{b_i\}_{i=1}^{N_{\rm lat}}$ 
            from $\big\{\wh{\GG}_i \big\}_{i=1}^{N_{\rm lat}}$
            \vspace{1mm}
            \For{$i$ in $\{1,2,\cdots,N_{\rm lat}\}$}
                \State $F_i \leftarrow f_i + \beta b_i$
                \State Append the information of $F_i$ to $\tau_i$ 
                \Statex \qquad \quad in $D$
            \EndFor
            \State Update $\phi$ and $\psi$ via stochastic gradient
            \Statex \quad ~ descent 
            of loss\vspace{1mm}%
            \Statex
            \qquad $\displaystyle \mathcal{L}_{Q}(\phi) = 
            \sum_{(\ss, \aa, F_i) \in B}
            \mkakko{
                Q_{\phi}(\ss,\aa) - F_i
            }^2 $
            \Statex 
            \qquad $\displaystyle \mathcal{L}_{V}(\psi) = 
            \sum_{(\ss, F_i)\in B}
            \mkakko{
                V_{\psi}(\ss) - F_i
            }^2 $
            \vspace{1mm}
            \Statex \quad ~ 
            where a minibatch $B$ is drawn from $D$
            \State Compute corrected scores
            \vspace{1mm}
            \Statex 
            \qquad \qquad 
            $\wh{F}_i (\ss,\aa) = Q_\phi(\ss,\aa) - V_\psi(\ss)$
            \State Update $\theta$ via gradient descent of loss 
            \Statex 
            \quad $\displaystyle \mathcal{L}_{\pi}(\theta) =
            - \sum_{i=1}^{N_{\rm lat}}
            \sum_{(\ss,\aa)\in \tau_i} \wh{F}_i (\ss, \aa)
            \log \pi_{\theta}(\aa \vert \ss,\c_i) $
        \EndWhile
        \\
        \Return $\pi_\theta$
    \end{algorithmic}
\end{algorithm}
In this method we train two additional NN, $\{Q_\phi, V_\psi \}$, which we call \emph{generalized value functions} despite that they have nothing to do with the standard value functions in single-objective RL. The role of these NN is to estimate the contribution of individual states and actions within each trajectory to the full score $\{F_i\}_i$. Unlike in \NAME{}, score thresholding above zero is not required in \NAMEV{}: we simply have $F_i=f_i+\beta b_i$. Instability of learning is avoided thanks to the transition-wise update capability of \NAMEV{}. An important point to note is that $Q_\phi$ and $V_\psi$ \emph{do not depend on the latent variables} $\{\c_i\}_i$.

Both \NAME{} and \NAMEV{} require collecting $N_{\rm lat}$ trajectories at every iteration. This step can be accelerated efficiently by means of parallelization over multiple CPU cores.

\section{Environments\label{sc:gdf8g7wzz}}
In this section, we describe environments that are used to numerically evaluate the proposed algorithm in Sec.~\ref{sc:453rewfd}.

\subsection{Deep Sea Treasure}

\begin{figure}[t]
	\centering
	\includegraphics[width=.6\columnwidth]{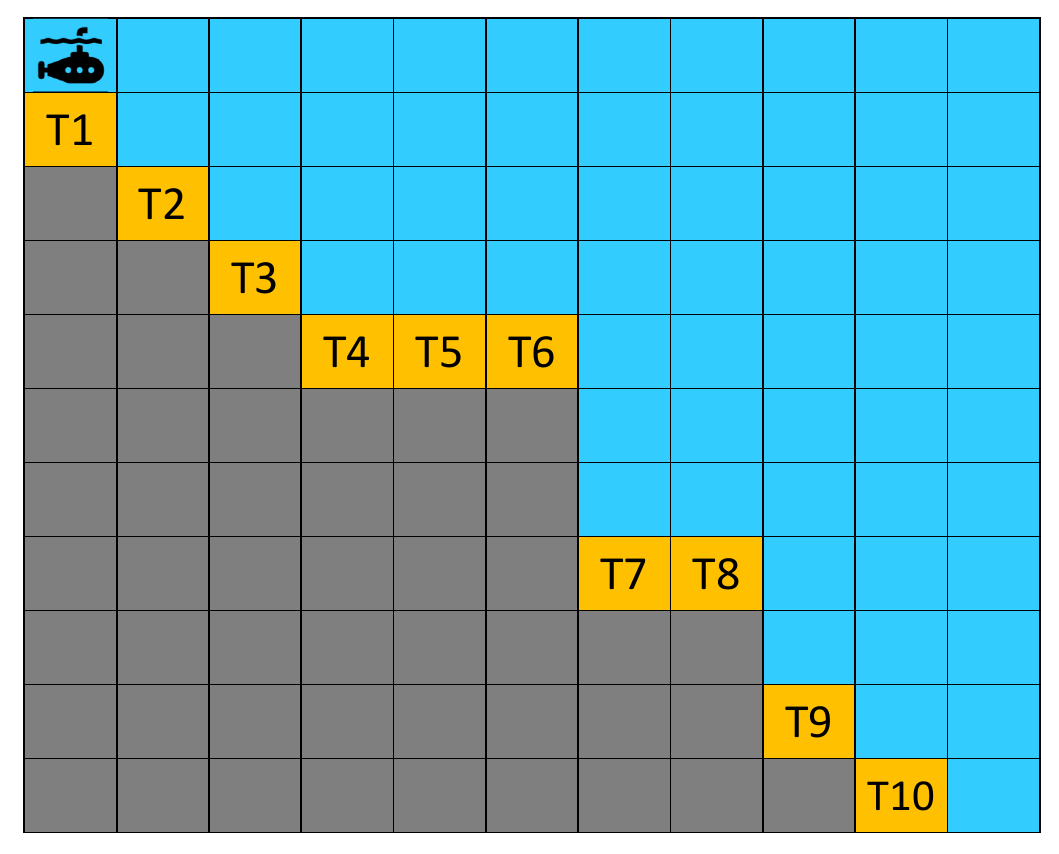}
	\caption{\label{fg_DST}The DST environment. Orange cells are treasures and blue cells are the ocean.}
\end{figure}

Deep Sea Treasure (DST) is a simple grid world proposed by Vamplew et al.~\cite{DBLP:journals/ml/VamplewDBID11}. Many prior works employed DST to evaluate MORL methods \cite{DBLP:journals/jmlr/MoffaertN14,DBLP:conf/ijcnn/ParisiPSBR14,Reymond2019,DBLP:conf/icml/AbelsRLNS19,DBLP:conf/nips/YangSN19,DBLP:conf/icml/AbdolmalekiHNS20,DBLP:conf/atal/ReymondBN22,DBLP:journals/corr/abs-2208-07914}. As shown in Fig.~\ref{fg_DST}, a submarine controlled by the agent starts from the top left cell and moves over the $11\times 11$ grid by taking one of the four actions \{up, down, left, right\}. There are two objectives in DST: time cost and treasure reward. At each time step, the agent incurs a time cost of $-1$. The treasure reward is $0$ if the agent is in blue cells and is equal to the treasure value $\in\{\T 1,\T 2,\cdots,\T 10\}$ if it reaches one of the treasures (orange cells). The episode ends if either the agent reaches a treasure or moves into a cliff (gray cells). Fortunately the exact PF of DST is known; it is given by trajectories of minimum Manhattan distance from the start cell to each treasure. 

There are some variations in the literature regarding the treasure values. In this work, we consider two settings.
\begin{itemize}
	\item Original version \cite{DBLP:journals/ml/VamplewDBID11,DBLP:journals/jmlr/MoffaertN14,DBLP:conf/ijcnn/ParisiPSBR14,Reymond2019,DBLP:conf/atal/ReymondBN22}:~$\{{\T}n\}_{n=1}^{10}=$ \\
	$(1,2,3,5,8,16,24,50,74,124)$.
	\item Convex version \cite{DBLP:conf/nips/YangSN19,DBLP:conf/icml/AbdolmalekiHNS20,DBLP:journals/corr/abs-2208-07914}:~$\{{\T}n\}_{n=1}^{10}=$ \\
	$(0.7, 8.2, 11.5, 14.0, 15.1, 16.1, 19.6, 20.3, 22.4, 23.7)$.
\end{itemize}
The convex version is easier to solve because all solutions on the PF can in principle be discovered with linear scalarization methods. 
Note that upper treasures such as \T1 and \T2 are easy to find while farther treasures such as \T10 are substantially more difficult to find. Thus, despite its deceptive simplicity, DST poses a hard challenge of solving the exploration vs.~exploitation dilemma in RL.

\subsection{Fruit Tree Navigation}

Fruit Tree Navigation (FTN) \cite{DBLP:conf/nips/YangSN19} is a binary tree of depth $d$. The state space is discrete and two-dimensional: $\ss=(i,j)\in\NN^2$ with $0\leq i \leq d$ and $0 \leq j \leq 2^i-1$. At every non-terminal node, the agent selects between left and right. At a terminal node the agent receives a reward $\rr\in\RR^6$ and finishes the episode. Thus the length of every episode is equal to $d$. The set of reward values is configured such that \emph{every reward is on the PF}. Now the challenge for MORL methods is to discover and maintain all the $2^d$ Pareto-optimal policies, which gets harder for higher $d$. In this paper we consider $d\in\{5,6,7\}$ following \cite{DBLP:conf/nips/YangSN19,DBLP:journals/corr/abs-2208-07914}. We use the code available at GitHub \cite{yanggithub2019}.

\subsection{Linear Quadratic Gaussian Control\label{sc:myz834}}

Linear Quadratic Gaussian Control (LQG) is a well-known classic problem in control theory \cite{10.5555/561399} with multi-dimensional continuous state and action spaces. LQG has been considered as a test problem in a number of RL and MORL studies, e.g., \cite{DBLP:journals/nn/PetersS08,DBLP:conf/ijcnn/ParisiPSBR14,DBLP:journals/jair/ParisiPR16,DBLP:journals/ijon/ParisiPP17,DBLP:conf/icml/Fazel0KM18}. Specifically, in this paper we consider the multi-objective version presented in \cite{DBLP:conf/ijcnn/ParisiPSBR14,DBLP:journals/jair/ParisiPR16,DBLP:journals/ijon/ParisiPP17}. Let $m$ denote the dimension of state and action spaces. The state transition dynamics is defined by
\begin{align}
	\ss_{t+1} = \ss_{t} + \aa_{t} + \sigma {\bm\varepsilon}\,,
	\label{eq:egdv43}
\end{align}
where $\ss\in\RR^m$ and $\aa\in\RR^m$ are state and action, ${\bm\varepsilon}$ is a random noise drawn from the $m$-dimensional normal distribution, and \mbox{$\sigma\geq 0$} is the noise strength parameter. The reward $\rr\in\RR^m$ is defined by
\begin{align}
	r_i = - \ss^\T Q_i \ss - \aa^\T R_i \aa \qquad (i=1,...,m)
	\label{eq:4lkhlkfdus09}
\end{align}
where $Q_i$ and $R_i$ are $m\times m$ diagonal matrices. All diagonal entries of $Q_i$ are equal to $\xi$ except $(Q_i)_{ii}=1-\xi$, whereas all diagonal entries of $R_i$ are equal to $1-\xi$ except $(R_i)_{ii}=\xi$. Following \cite{DBLP:conf/ijcnn/ParisiPSBR14,DBLP:journals/jair/ParisiPR16,DBLP:journals/ijon/ParisiPP17}, we set $\xi=0.1$ and the initial state to $\ss_0=(10,...,10)^\T$. As for the noise we consider both $\sigma=0$ and $1$. 

Part of the motivation to consider LQG for benchmarking stems from the fact that the optimal policy for LQG is known \cite{10.5555/561399,Tedrake2023}. For simplicity, let us consider a single-objective problem with scalar reward $r=-\ss^\T Q \ss - \aa^\T R \aa$ and dynamics \eqref{eq:egdv43}, where $Q$ and $R$ are $m\times m$ positive-definite matrices. Then the optimal control that maximizes the discounted cumulative rewards $\sum_{t=0}^{\infty}\gamma^t r_t$ is given by 
\begin{align}
	\aa = - \gamma (R + \gamma S)^{-1}S \ss \,.
	\label{eq:43rfdsf}
\end{align}
The positive-definite square matrix $S$ is obtained by numerically solving the discrete-time Riccati difference equation
\begin{align}
	S = Q + \gamma S - \gamma^2 S (R + \gamma S)^{-1} S\,.
	\label{eq:opi2349}
\end{align}
Note that \eqref{eq:43rfdsf} and \eqref{eq:opi2349} do not depend on $\sigma$. Thus, under the assumption that the PF for the multi-objective problem \eqref{eq:4lkhlkfdus09} is convex, the entire Pareto set can be obtained by iteratively solving \eqref{eq:43rfdsf} and \eqref{eq:opi2349} using the linearly weighted sums $Q=\sum_{i=1}^{m}w_i Q_i$ and $R=\sum_{i=1}^{m}w_i R_i$ for a sufficiently dense set of weights $\mathbf{w}\in\RR^m$, $w_i\geq 0$ $\forall i$ and $\sum_{i=1}^{m}w_i=1$. The convexity of the PF has been verified in \cite{DBLP:conf/ijcnn/ParisiPSBR14,DBLP:journals/jair/ParisiPR16,DBLP:journals/ijon/ParisiPP17}.

\subsection{Minecart}

\begin{figure}[t]
	\centering
	\includegraphics[width=.6\columnwidth]{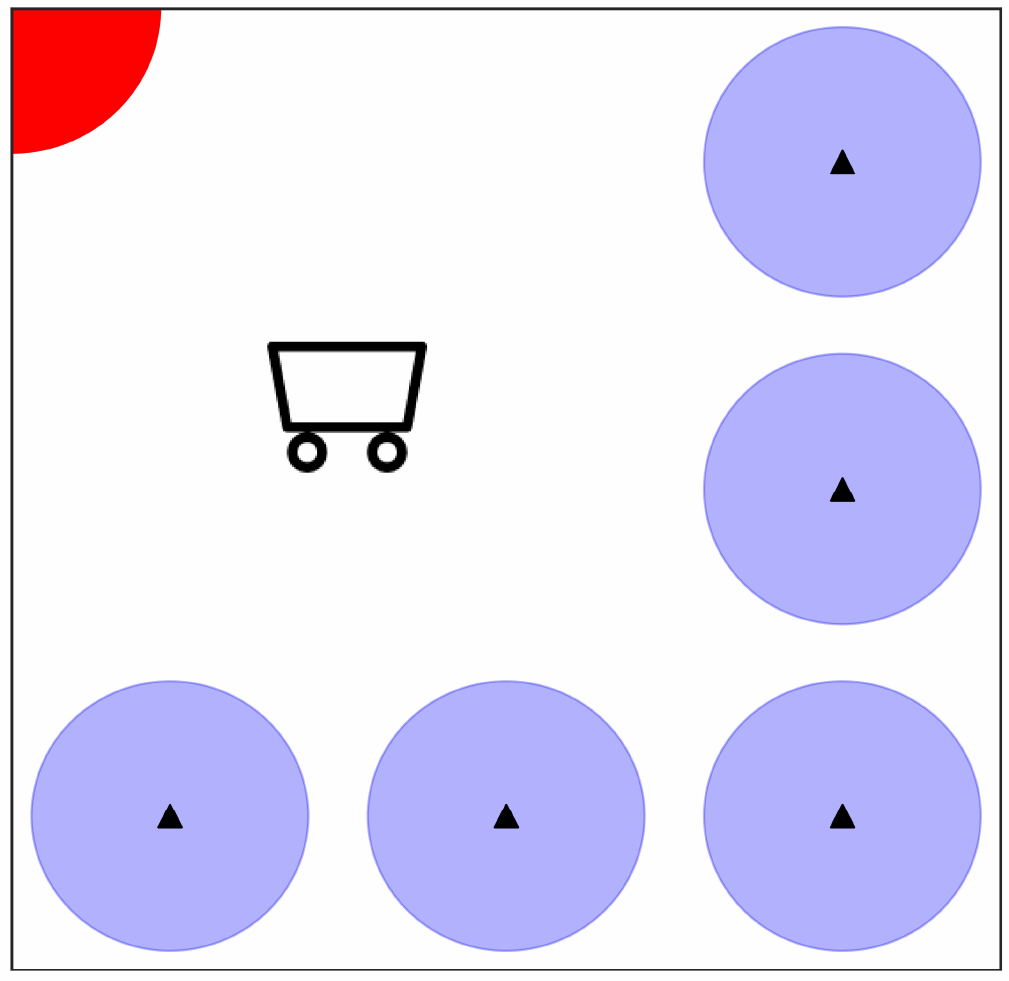}
	\caption{\label{fg_minecart_env}Minecart environment. The cart departs from the top-left corner, goes for mining at any of the 5 mines (blue circles), and returns home (red quarter circle) to sell ores.}
\end{figure}

Minecart \cite{DBLP:conf/icml/AbelsRLNS19} is a multi-objective environment with a continuous state space, designed for testing MORL algorithms. The map of Minecart is shown in Fig.~\ref{fg_minecart_env}. 
The cart is initially at the top-left corner $(0,0)$ and can go anywhere inside the unit square $[0,1]^2$. There are five mines, to which the cart must travel to perform mining. Inside a mine (blue circle) the cart can mine and get two kinds of ores. The amount of ores available via one-time mining is different for each mine. The cart's capacity is $1.5$ so that the sum of mined ores on the cart cannot exceed this capacity. Once the cart returns to the home port (red circle), it sells all the ores, acquires rewards, and terminates the episode. The return in Minecart is three-dimensional: the first two components are the amount of mined ores, and the last component is the total fuel cost.  The cart can take six discrete actions, [\emph{Mine}, \emph{Turn Left}, \emph{Turn Right}, \emph{Accelerate}, \emph{Brake}, \emph{Do Nothing}]. The underlying state space is $\RR^6$ consisting of the cart's position, speed, angle, the mined amounts of ore1 and ore2. 

To get ores while minimizing the fuel cost, the cart must go straight to the desired mine(s), do mining, rotate the direction, and return straight to the home port. This requires a highly nonlinear control and offers an attractive testbed for deep MORL approaches. We use the source code of Minecart available at \cite{Reymond_github}.

\section{Numerical experiments\label{sc:453rewfd}}

In this section we conduct several experiments to evaluate our method. 

\subsection{Implementation details}
We use PyTorch to implement all NN models. Initial parameters of NN are randomly drawn from $\mathcal{N}(0, 0.2^2)$. NN are optimized with ADAM stochastic optimizer \cite{DBLP:journals/corr/KingmaB14} with learning rate 1e$-$3. Tanh activation is used for the first activation of the latent input to policy NN. SELU activation \cite{DBLP:journals/corr/KlambauerUMH17} is used for all the other activation layers in policy NN, $Q_\phi$, and $V_\psi$. The computation of HV is done with external multi-objective optimization library \textsf{pymoo} \cite{pymoo}. We use an Intel Xeon Gold 5218 CPU with 2.30GHz and an NVIDIA Tesla V100 GPU for our experiments. 

Unless stated otherwise, numerical plots in this section show the mean $\pm$ one standard deviation over 5 independent runs with different random seeds. 
In learning curve plots, the abscissa \emph{iteration} refers to the number of gradient updates performed on the policy.
During training, quality of the PF produced by the policy (conditioned on $N_{\rm lat}$ latent variables $\{\c_i\}$ randomly drawn from $P(\c)$) is evaluated on test episodes at every iteration round. The number of test episodes per iteration is 1 for deterministic environments and $>1$ for stochastic environments, for which the average of scores over multiple test episodes is recorded. The mean and the standard deviation of the best test scores from 5 training runs are reported in tables for comparison with baselines.

\subsection{Results for DST}

As noted previously, there are two settings for DST environment. The true PF for both cases are displayed in Fig.~\ref{fg_dst_pf}. Hyperparameters used for the experiment are summarized in Appendix~\ref{sc:4332}. 
\begin{figure}[t]
	\centering
	\includegraphics[width=.9\columnwidth]{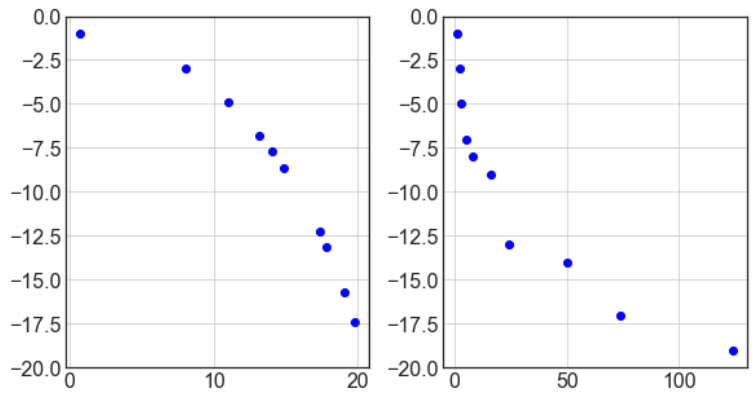}
	\caption{\label{fg_dst_pf}Exact PF of the DST environment. The vertical axis is the cumulative time cost, and the horizontal axis is the treasure reward. {\bf Left:} For convex treasure values and $\gamma=0.99$. {\bf Right:} For original treasure values and $\gamma=1.0$. These discount rates were chosen to ensure a fair comparison with related work.}
\end{figure}

In the setting with convex PF, our \NAME{} yielded the result in Table~\ref{tb:dstconv}. To compute HV we have used $(0, -19)$ as a reference point.
\begin{table}[tb]
	\centering
	\caption{\label{tb:dstconv}Comparison of methods in the DST (convex PF) environment. Results for baselines are taken from \cite{DBLP:conf/nips/YangSN19,DBLP:journals/corr/abs-2208-07914}. Best values are marked with an asterisk.}
	\begin{tabular}{lcc}
		\toprule
		& {\bf CRF1} ($\uparrow$) & {\bf HV} ($\uparrow$) 
		\\\midrule 
		{\bf Envelope} \cite{DBLP:conf/nips/YangSN19} & 0.994 & 227.89
		\\
		{\bf CN$+$DER} \cite{DBLP:conf/icml/AbelsRLNS19} & 0.989 & ---
		\\
		{\bf PD-MORL} \cite{DBLP:journals/corr/abs-2208-07914} & 1.0$^*$ & 
		$241.73^*$
		\\
		{\bf \NAME{} (Ours)} & 1.0$^*$ & 241.73 $\pm 0^*$
		\\\bottomrule
	\end{tabular}
\end{table}
The score CRF1 (higher is better, the maximum is 1), which was defined in \cite{DBLP:conf/nips/YangSN19}, is also shown for reference. We find that \NAME{} found the true PF for all runs. The bonus factor plays an essential role in obtaining this result. 
\begin{figure}[t]
	\centering
	\includegraphics[width=.6\columnwidth]{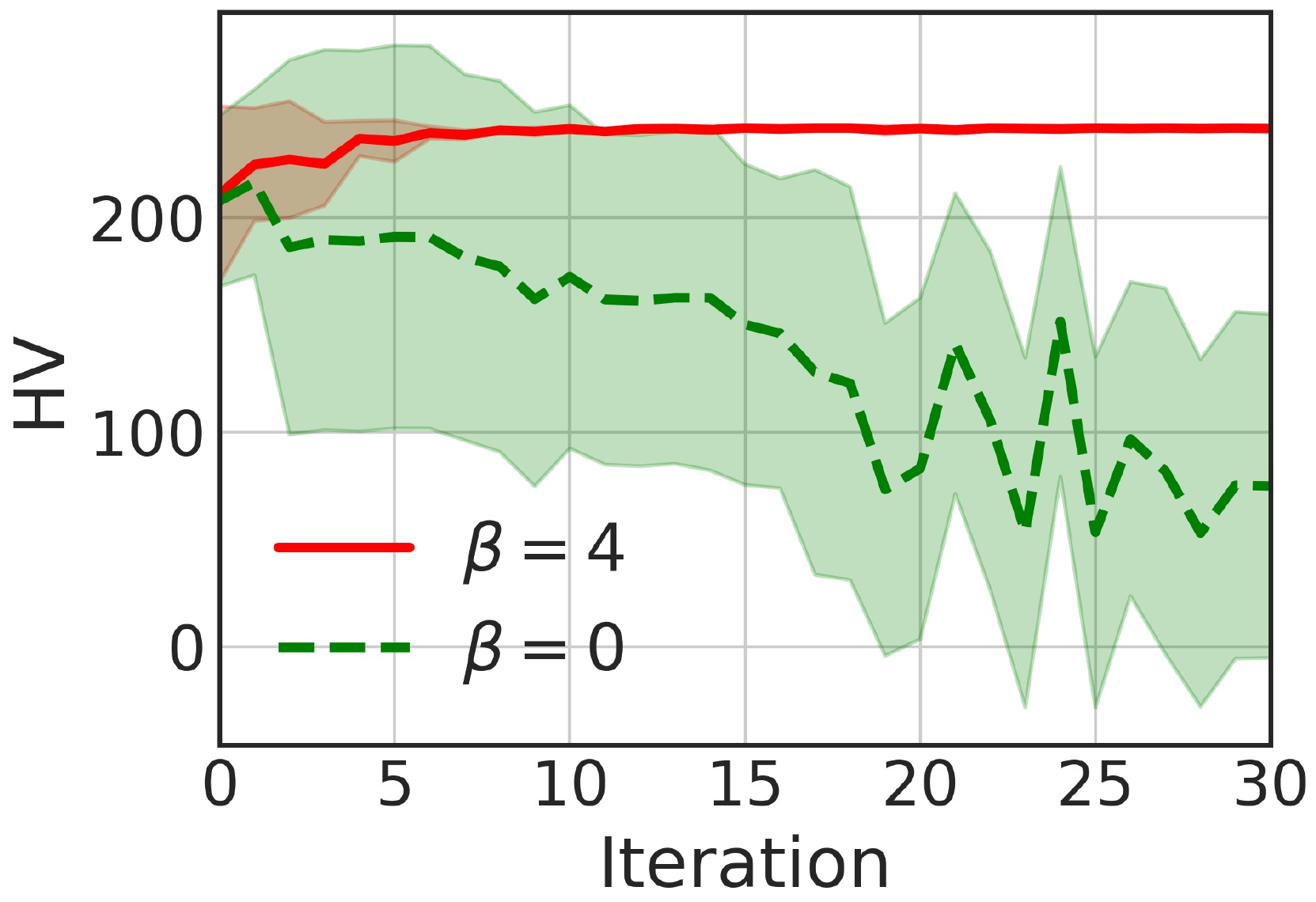}
	\vspace{1mm}\\
	\hspace{-4mm}\includegraphics[width=.65\columnwidth]{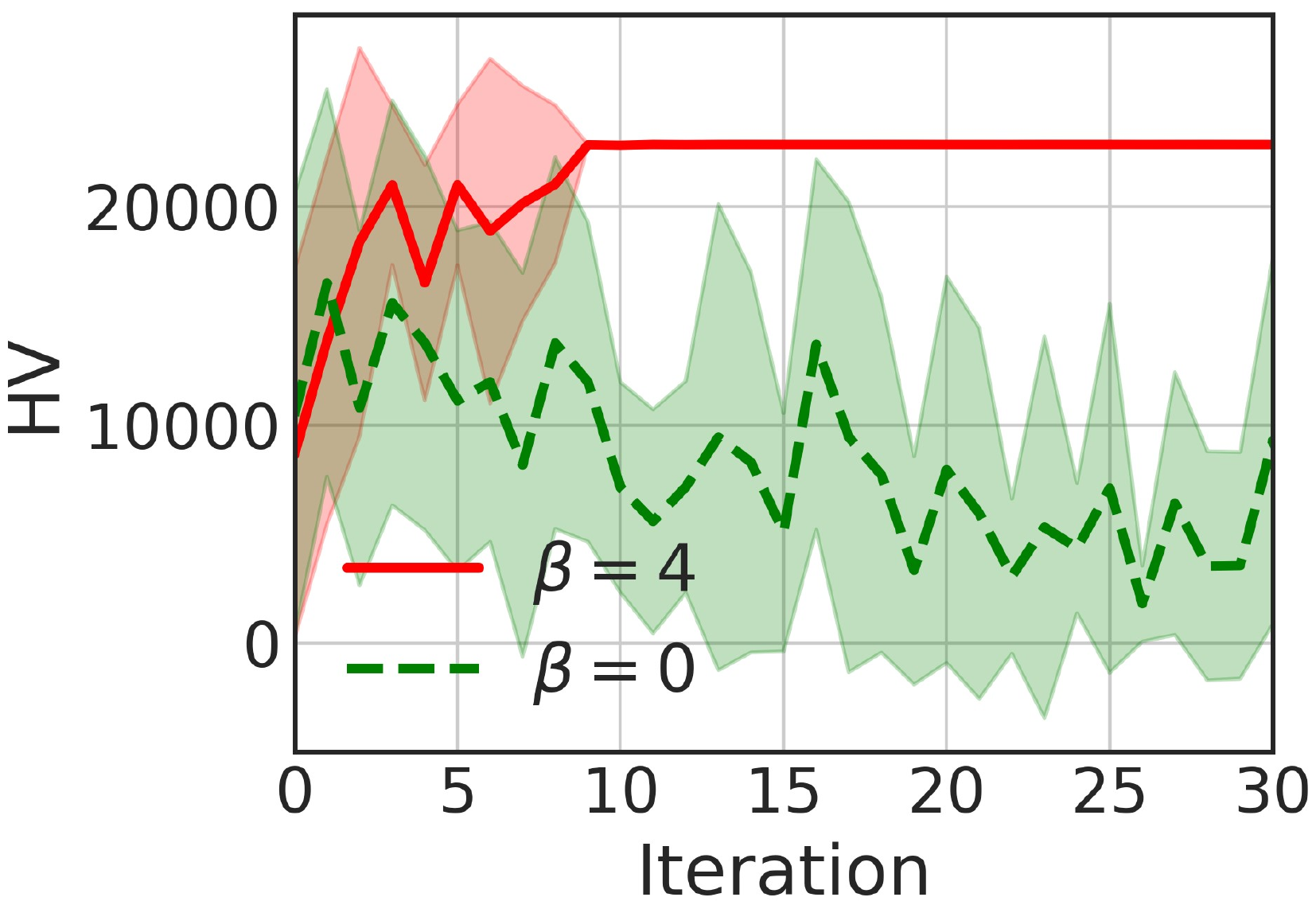}
	\vspace{2mm}
	\caption{\label{fg_learning_dst}Training process of \NAME{} for the DST environment. {\bf Top:} For convex treasure values. {\bf Bottom:} For original treasure values.}
\end{figure}
In Fig.~\ref{fg_learning_dst} (top) we plot the learning curves of \NAME{} with and without bonus. We observe that, without bonus, the policy is trapped in local optima (i.e., the treasures \T1 and \T2 located near the initial position) and never discovers the true PF. 

The results for DST with the original treasure values are presented in Table~\ref{tb:dstorg}. To compute HV we have used $(0, -200)$ as a reference point.%
\begin{table}[tb]
	\centering
	\caption{\label{tb:dstorg}Comparison of methods in the DST (original) environment. Results for baselines are taken from \cite{DBLP:conf/atal/ReymondBN22}. Best value is marked with an asterisk.}
	\begin{tabular}{lc}
		\toprule
		& {\bf HV} $(\uparrow)$
		\\\midrule
		{\bf RA} \cite{DBLP:conf/ijcnn/ParisiPSBR14} & 
		$22437.40 \pm 49.20$
		\\
		{\bf MO-NES} \cite{DBLP:journals/ijon/ParisiPP17} & 
		$17384.83 \pm 6521.10$
		\\
		{\bf PCN} \cite{DBLP:conf/atal/ReymondBN22} & 
		$22845.40 \pm 19.20$
		\\
		{\bf \NAME{} (Ours)} & $22855.0 \pm 0^*$
		\\\bottomrule 
	\end{tabular}
\end{table}
Unlike other methods, \NAME{} has successfully found the true PF for all runs. This result may not be obtained when the bonus is zero, as shown in the learning curves in Fig.~\ref{fg_learning_dst} (bottom).

\subsection{Results for FTN}

Next, we report numerical results obtained with \NAME{} in the FTN environment with the depth parameter $d=5,6,7$. The true PF comprises $2^d$ points and the problem becomes harder for higher $d$. Hyperparameters used in \NAME{} are summarized in Appendices \ref{sc:fktrdfg89}, \ref{sc:fkerw9} and \ref{sc:8ewr39}. We have utilized state embedding in this environment. Namely, a state $\ss=(i,j)\in\NN^2$ is first normalized as $(i/d, j/2^i)$ and then embedded into $\RR^{e1}\times \RR^{e2}$ via cosine embedding \eqref{eq:4mbpwqf}. The values $(e1, e2)\in\NN^2$ are listed in the appendices. 

The performance of our method is compared with baselines in Table~\ref{tb:ftn}.
\begin{table*}[tb]
	\centering
	\caption{\label{tb:ftn}Comparison of methods in the FTN environment. Results for baselines are taken from \cite{DBLP:journals/corr/abs-2208-07914}. Best values are marked with an asterisk.}
	\scalebox{0.84}{
	\begin{tabular}{lccc}
		\toprule
		& \multicolumn{3}{c}{{\bf HV} $(\uparrow)$}
		\\\cmidrule(){2-4}
		& $d=5$ & $d=6$ & $d=7$
		\\\midrule
		{\bf Envelope} \cite{DBLP:conf/nips/YangSN19} 
		& 6920.58$^*$ & 8427.51 & 6395.27
		\\
		{\bf PD-MORL} \cite{DBLP:journals/corr/abs-2208-07914} 
		& 6920.58$^*$ & 9299.15 & 11419.58
		\\
		{\bf \NAME{} (Ours)} 
		& 6920.58 $\pm$ 0$^*$ & 9302.38 $\pm$ 0$^*$ & 
		12290.93 $\pm$ 22.82$^*$
		\\\bottomrule 
	\end{tabular}}
\end{table*}
The origin $(0,\cdots,0)\in\RR^6$ was used as a reference point for HV.  
For $d=5$ and $6$, our method was able to discover the true PF in all runs. For $d=7$ our method discovered the true PF in 4 out of 5 runs. The mean score is hence very close to the highest possible HV for $d=7$, which is $12302.34$. We conclude that \NAME{} solved all cases almost exactly. For $d=7$ our score is $7.6\%$ higher than the best baseline score (PD-MORL). However, we found that very careful tuning of the bonus parameters $\{k, \beta\}$, the latent dimension $d_{\rm lat}$, and the hidden layer width of policy NN was necessary to attain this result. 

\begin{figure}[t]
	\centering
	\includegraphics[width=.47\columnwidth]{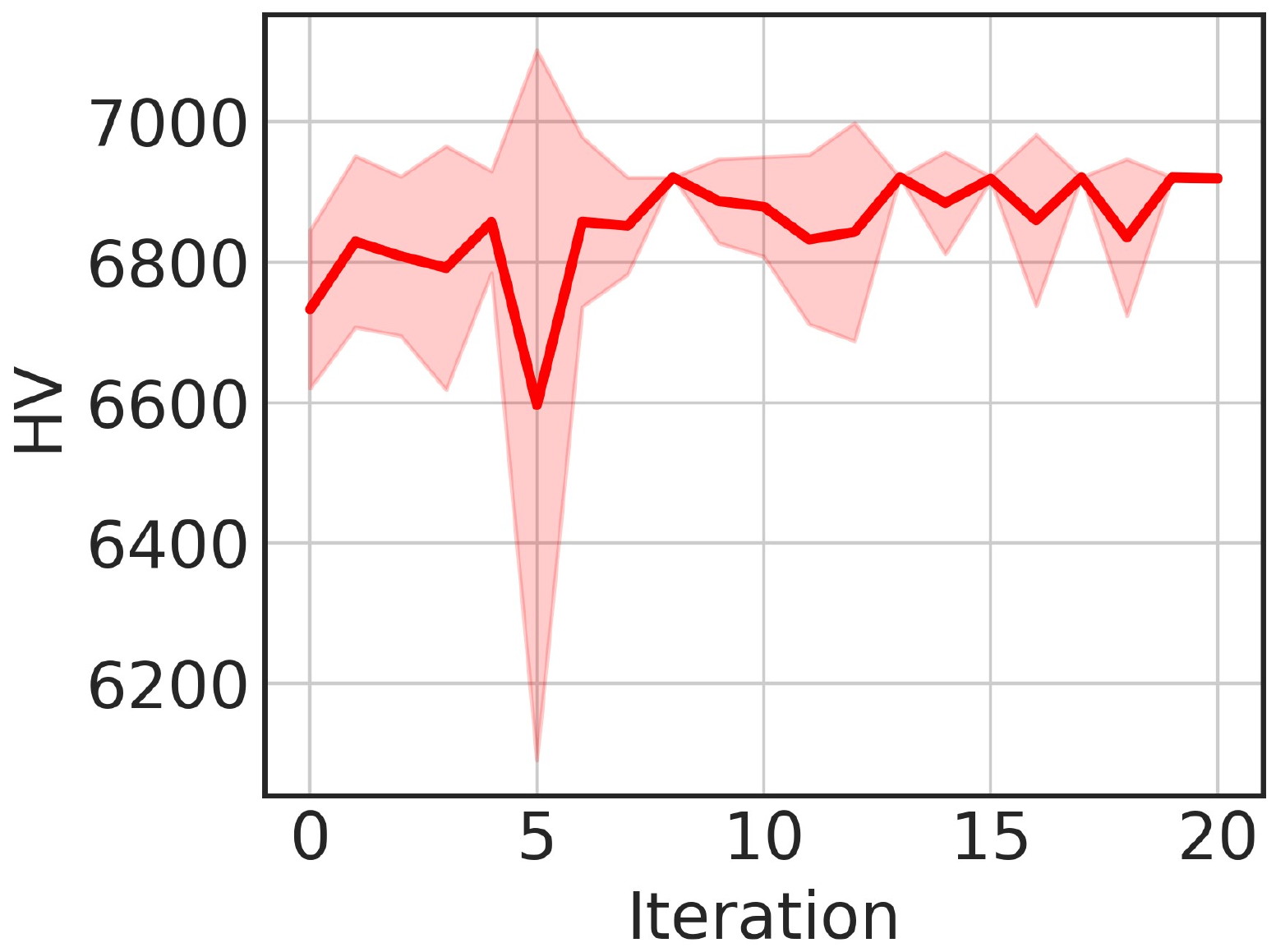}
	\includegraphics[width=.47\columnwidth]{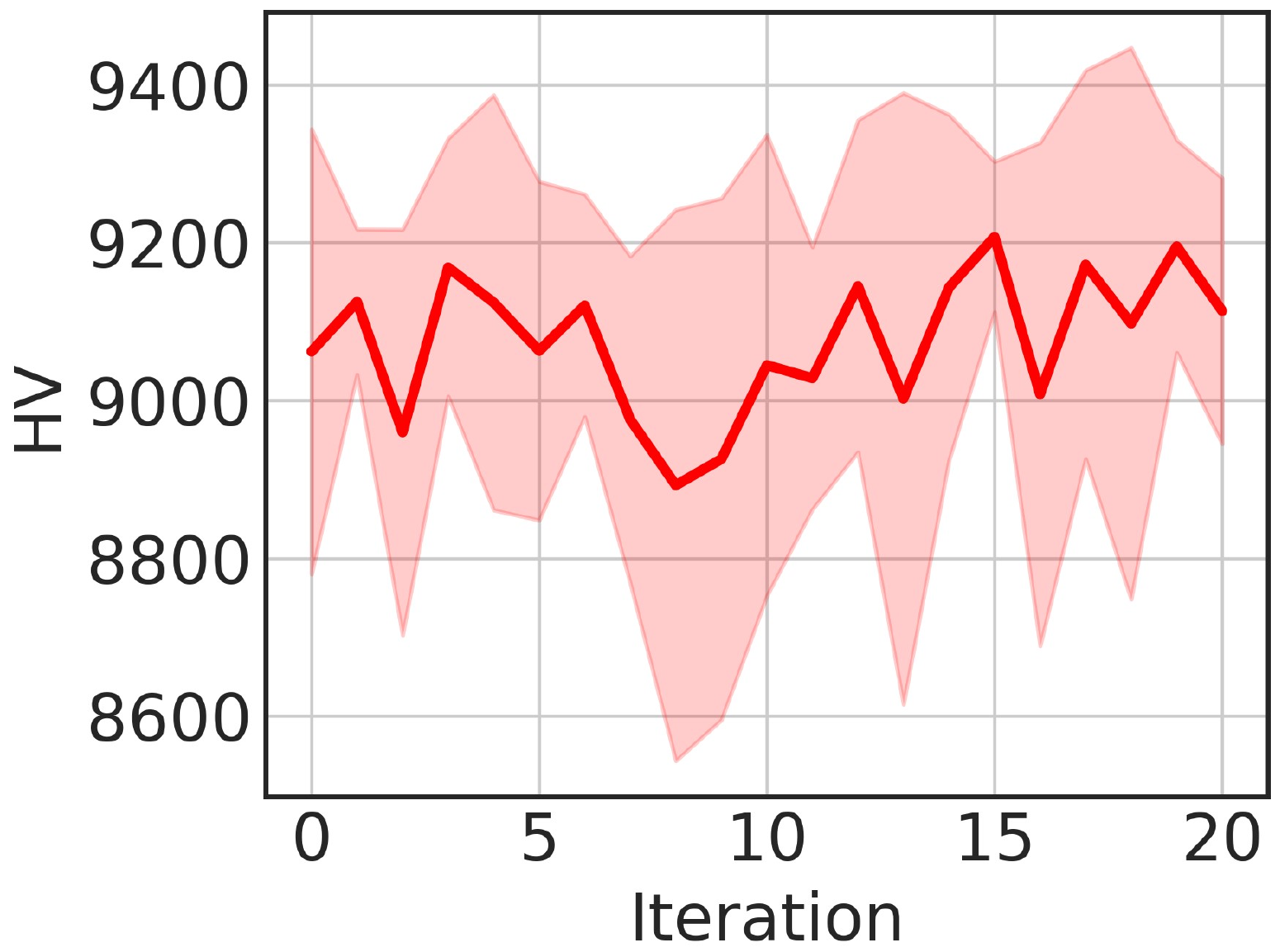}
	\put(-150,20){$d=5$}
	\put(-30,20){$d=6$}
	\vspace{3mm}
	\\
	\includegraphics[width=.47\columnwidth]{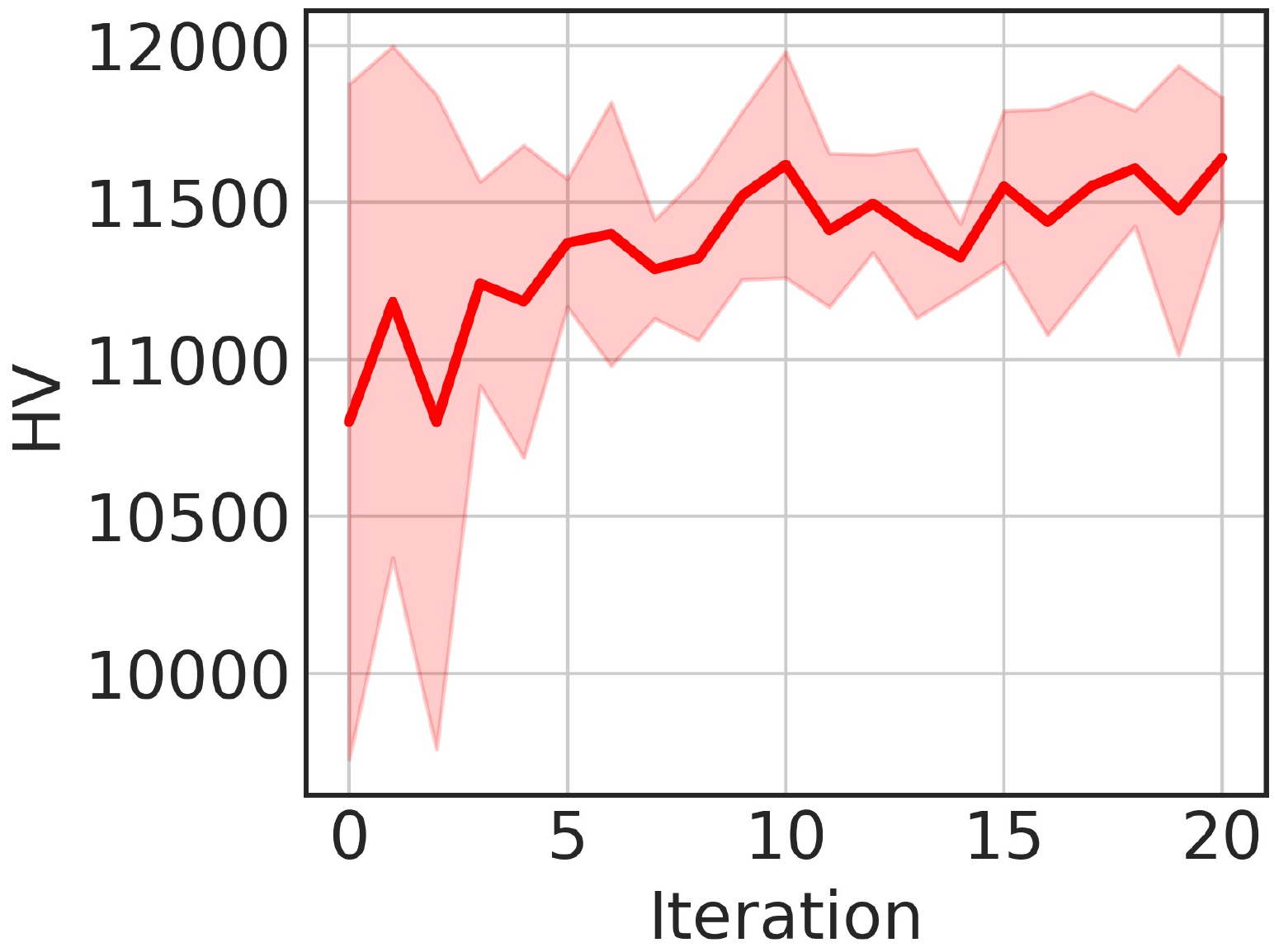}
	\put(-32,20){$d=7$}
	\vspace{2mm}
	\caption{\label{fg_learning_ftn}Training process of \NAME{} for the FTN environment with depth $d=5,6,7$.}
\end{figure}

The learning curves of \NAME{} are displayed in Fig.~\ref{fg_learning_ftn}. 
As the computation of HV becomes exponentially hard in high dimensions, 
we monitored the performance at every iteration using $N_{\rm lat}=(300,400,400)$ for $d=5,6,7$, respectively, and measured the scores of the best policies using $N_{\rm lat}=1500$ in the evaluation phase. This explains why the scores in Table~\ref{tb:ftn} are higher than those in Fig.~\ref{fg_learning_ftn}. It must be noted that, although our method can parametrize infinitely many policies with a single NN, their performance evaluation can be computationally challenging especially when the number of objectives is high $(>3)$. In our experiment, the evaluation of 1500 policies took about half an hour.

\subsection{Results for LQG}

Next, we proceed to the evaluation of our method in the LQG environment. In this section we aim to: (i) Compare the PF of \NAME{} with the ground-truth PF. (ii) Visualize the evolution of the PF of \NAME{} during the training phase. (iii) Check dependence on the latent dimension $d_{\rm lat}$. (iv) Compare the performance of \NAME{} and \NAMEV{}. (v) Check if \NAME{} and \NAMEV{} work when the transition dynamics is stochastic. Hyperparameters used in the experiment are tabulated in Appendices~\ref{sc:lqg2} and \ref{sc:lqg3}. 

\paragraph{Two dimensions} 
The results for deterministic LQG in two dimensions $(\sigma=0)$ are summarized in Table~\ref{tb:pm45xvr96}.%
\begin{table}[tb]
	\centering
	\caption{\label{tb:pm45xvr96}Comparison of methods in the two-dimensional deterministic LQG environment $(\sigma=0)$.}
	\begin{tabular}{lc}
		\toprule
		& {\bf HV} $(\uparrow)$
		\\\midrule
		{\bf Optimal value} & 1.1646
		\\ 
		{\bf \NAME{} $(d_{\rm lat}=1)$} & 
		$1.1408 \pm 0.0061$
		\\
		{\bf \NAME{} $(d_{\rm lat}=2)$} &
		$1.1457 \pm 0.0040$
		\\
		{\bf \NAME{} $(d_{\rm lat}=3)$} &
		$1.1408 \pm 0.0076$
		\\
		{\bf \NAMEV{} $(d_{\rm lat}=2)$} &
		$1.1031 \pm 0.0090$
		\\\bottomrule 
	\end{tabular}
\end{table}
HV is computed with a reference point $(-310, -310)$ and divided by $160^2$. The optimal HV value is obtained with the method of Sec.~\ref{sc:myz834} using a mesh of weights $(w, 1-w)$ with $w\in\{0.01,0.02,\cdots,0.99\}$. It is observed that \NAME{} with $d_{\rm lat}=2$ attained the best score equal to 98.0\% of the true HV. We have also tested \NAMEV{} that uses the generalized value functions $Q$ and $V$, and obtained the final score equal to 94.7\% of the true HV. It is concluded that \NAME{} was able to solve the problem accurately and that \NAME{} outperformed \NAMEV{} on this problem. 

We have also investigated \NAME{} without clipping of scores above zero (line 11 of Algorithm~\ref{alg:main1}). It led to diverging gradient updates and policy parameters soon became NaN at an early stage of training. As a result, no sensible policy was obtained.

\begin{figure}[t]
	\centering
	\includegraphics[width=.6\columnwidth]{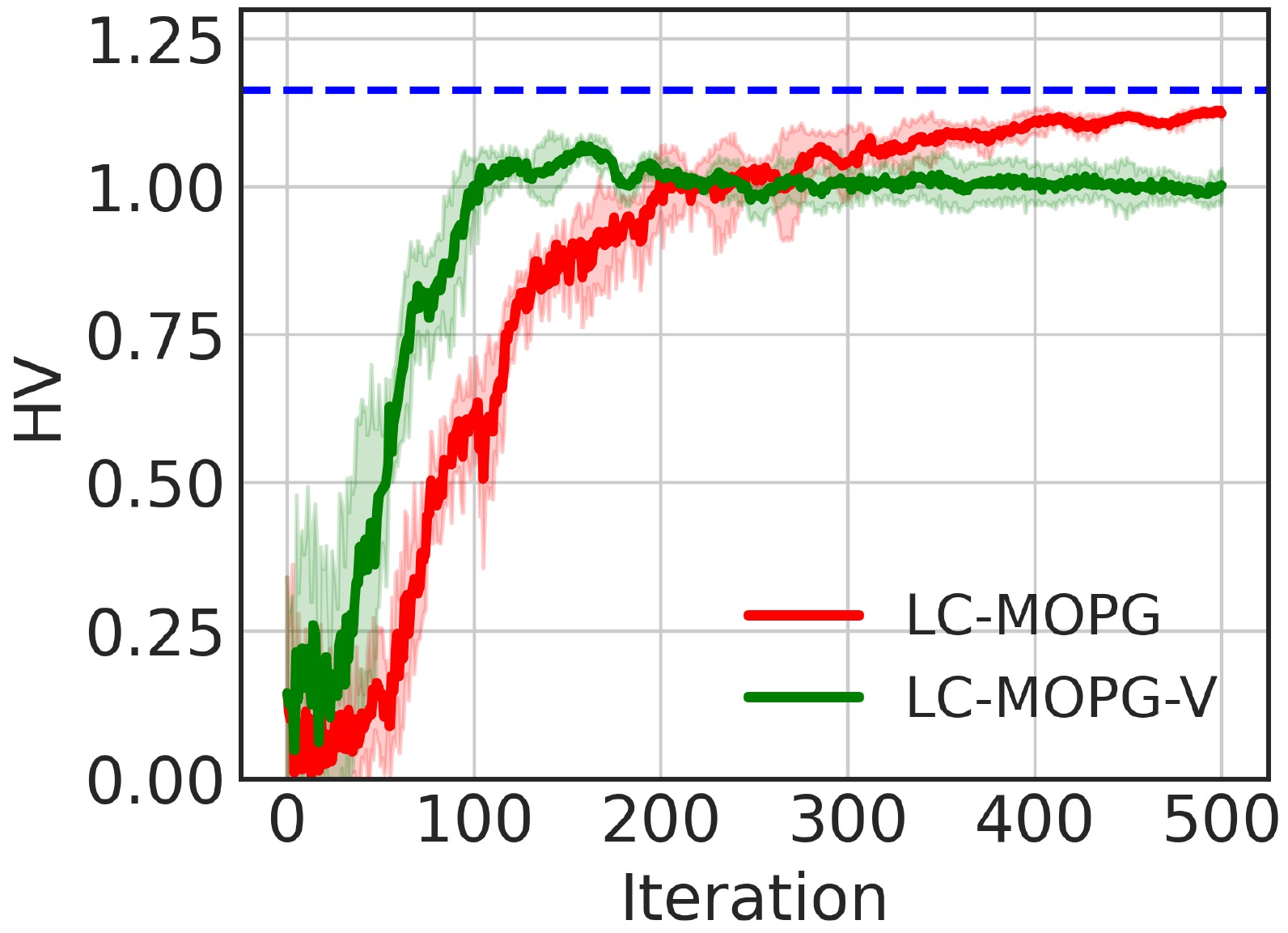}
	\caption{\label{fg_learning_lqg_2d}Comparison of the training process of our methods for two-dimensional deterministic LQG environment. The blue dashed line is the reference value obtained by solving the algebraic Riccati equation.}
\end{figure}

The learning curves of \NAME{} and \NAMEV{} (both for $d_{\rm lat}=2$) are compared in Fig.~\ref{fg_learning_lqg_2d}.  
It is observed that the initial learning of \NAMEV{} is roughly twice faster than that of \NAME{}, which may be attributed to the accurate score estimation of state-action pairs thanks to $Q$ and $V$. However, \NAMEV{}'s performance saturates early and is eventually surpassed by \NAME{} after 200 iterations. 

\begin{figure}[t]
	\centering
	\includegraphics[width=.99\columnwidth]{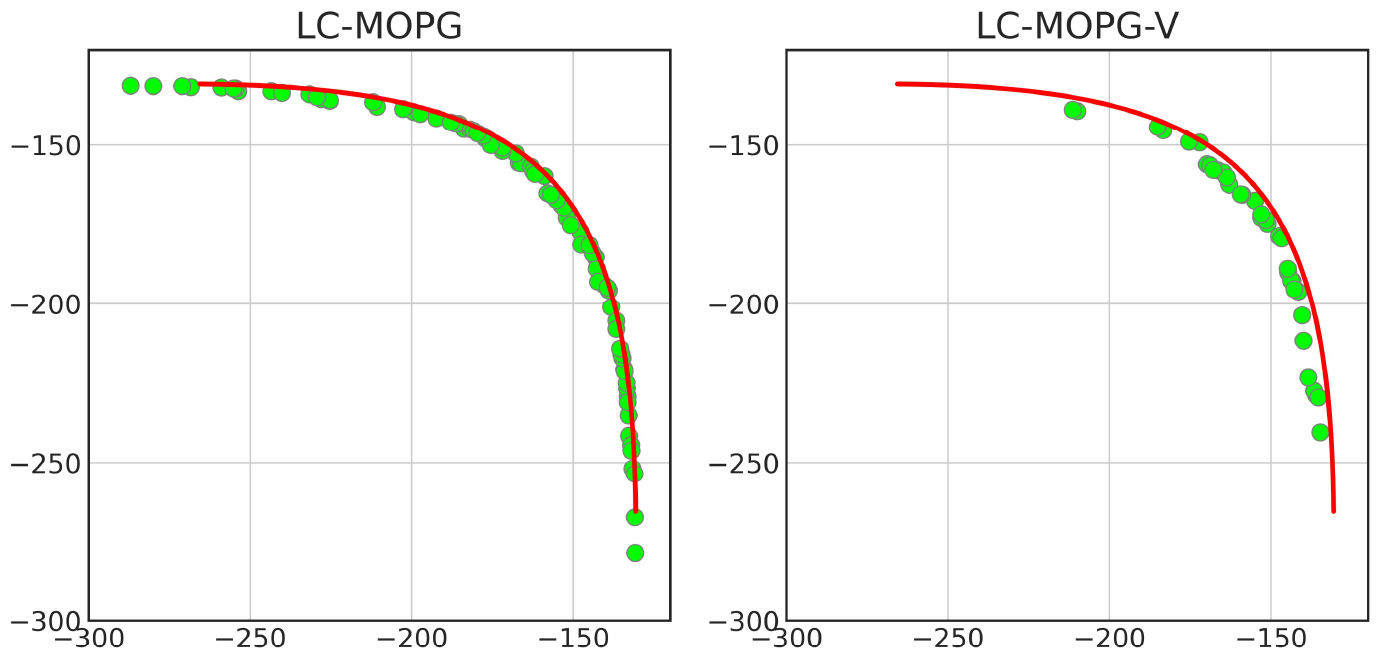}
	\vspace{2mm}
	\caption{\label{fg_pf_lqg}Comparison of the PF obtained with \NAME{} (left) and \NAMEV{} (right) in the two-dimensional LQG environment, overlayed with the optimal PF (red solid line) obtained by solving the algebraic Riccati difference equation.}
\end{figure}

To visually inspect their difference in the objective space, we randomly sampled 1000 latent variables, fed them into the trained policy networks of \NAME{} and \NAMEV{}, obtained their returns, and identified the PF. The result is displayed in Fig.~\ref{fg_pf_lqg}.%
For reference, the optimal PF obtained by solving the Riccati equation is also shown. The PF of both methods lie in the close vicinity of the true PF. However, the PF from \NAMEV{} is narrower than that of \NAME{}, implying that the issue with \NAMEV{} is coverage, rather than accuracy, of the PF. 

\begin{figure}[t]
	\centering
	\includegraphics[width=.97\columnwidth]{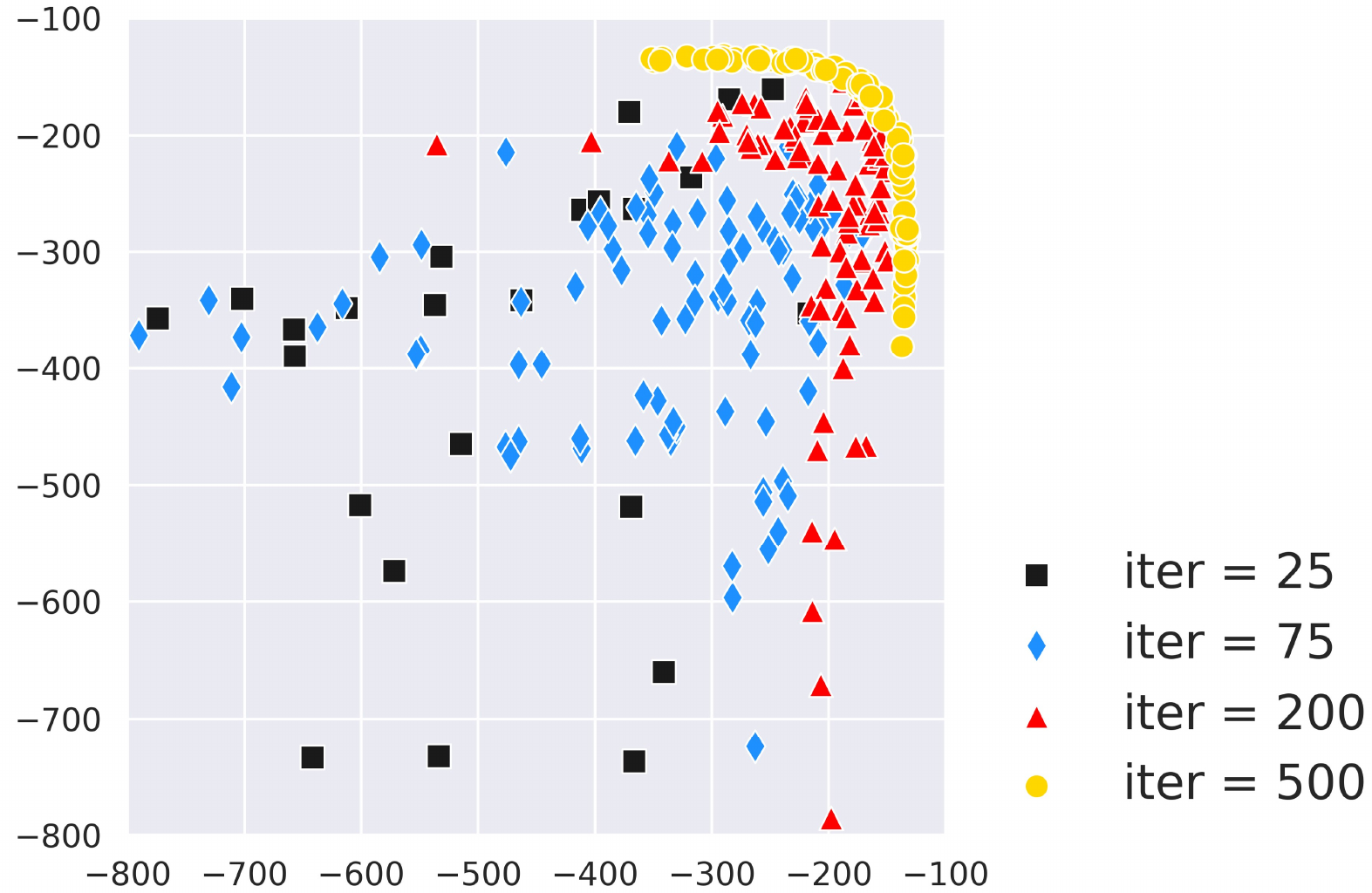}
	\vspace{2mm}
	\caption{\label{fg_lqg_evolve}Progression of the return set of \NAME{} during training in the two-dimensional LQG environment. At each iteration, 200 latent variables are randomly sampled, and the better half of returns (i.e., the set of $\GG_i$ for which $f_i > 0$ with $f_i$ computed as in Algorithm~\ref{alg:scoring}) is shown in the figure. Some low-lying returns are out of the scope of the figure.}
\end{figure}

The convergence of the return set $\{\GG_i\}_i$ towards the optimal PF during training is displayed in Fig.~\ref{fg_lqg_evolve}. 
One can see that the return set at iteration 25 is of low quality, but as the training progresses, the returns gradually move to the top-right area of the return space. At iteration 500 most returns cluster near the true PF. 

\begin{figure}[t]
	\centering
	\includegraphics[width=.6\columnwidth]{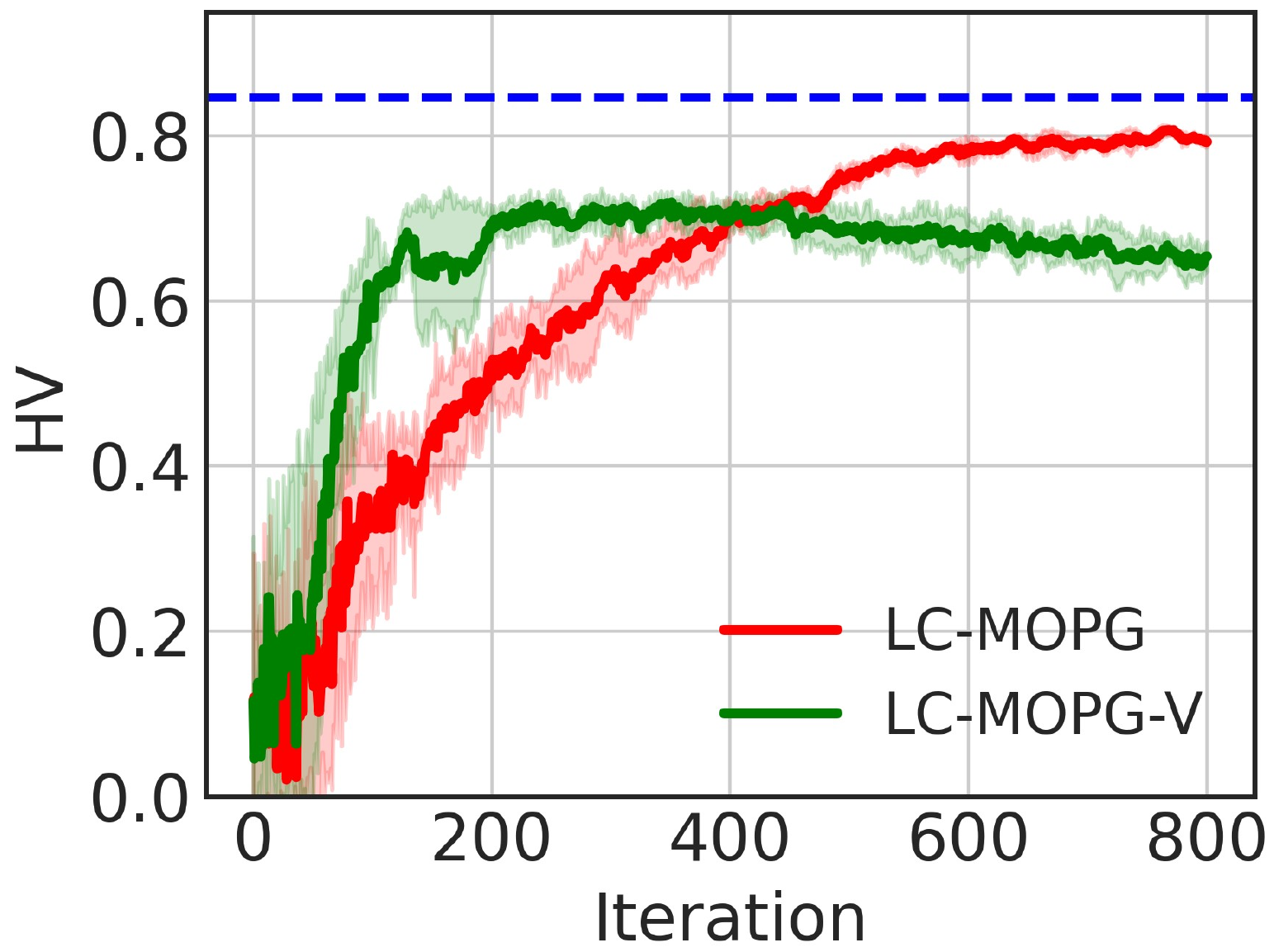}
	\caption{\label{fg_learning_lqg_3d}Same as Fig.~\ref{fg_learning_lqg_2d} but in three dimensions.}
\end{figure}

\paragraph{Three dimensions} 
Next, to see if our method works for more than two objectives, we evaluated the method in LQG in three dimensions without noise $(\sigma=0)$. The results are summarized in Table~\ref{tb:32gdlll}.  
\begin{table}[tb]
	\centering
	\caption{\label{tb:32gdlll}Comparison of methods in the three-dimensional deterministic LQG environment.}
	\begin{tabular}{lc}
		\toprule 
		& {\bf HV} $(\uparrow)$
		\\\midrule
		{\bf Optimal value} & 0.8476
		\\
		{\bf \NAME{} $(d_{\rm lat}=1)$} & $0.8124\pm 0.0036$
		\\
		{\bf \NAME{} $(d_{\rm lat}=2)$} & $0.8153\pm 0.0106$
		\\
		{\bf \NAME{} $(d_{\rm lat}=3)$} & $0.8208\pm 0.0037$
		\\
		{\bf \NAMEV{} $(d_{\rm lat}=3)$} &  
		$0.7544 \pm 0.0152$
		\\\bottomrule
	\end{tabular}
\end{table}
HV is computed with a reference point $(-500,-500,-500)$ and divided by $350^3$.  The optimal HV value is obtained with the method of Sec.~\ref{sc:myz834} using a mesh of 4851 weights \scalebox{0.9}{$\{(0.01, 0.01, 0.98), (0.01, 0.02, 0.97), \cdots, (0.98, 0.01, 0.01)\}$}. We observe that \NAME{} $(d_{\rm lat}=3)$ performs best, attaining HV equal to 96.8\% of the optimal value. It is surprising that \NAME{} $(d_{\rm lat}=1)$ has attained a high score despite the low latent dimension. We speculate that the latent space $[0,1]$ is windingly embedded in $\RR^3$ to densely approximate the two-dimensional PF. As for the impact of using generalized value functions, we once again find that \NAMEV{}'s performance is slightly inferior to that of \NAME{}, attaining HV equal to only 91.9\% of \NAME{}. Their learning curves are compared in Fig.~\ref{fg_learning_lqg_3d} (both for $d_{\rm lat}=3$). 
The qualitative behavior is the same as in two dimensions. Namely, \NAMEV{} achieves substantial speed-up of training compared to \NAME{} but quickly saturates at a suboptimal solution. It seems that the agent does not have enough time to randomly explore the environment, hence failing to find the global optimum and attracted to local optima. 
We have tried several tricks to fix this, including the use of different learning rates for the policy and value networks, and the clipping of trajectory scores above zero. However none of this produced qualitatively different results. How to cure this problem is left for future research. One direction could be to employ the generalized value functions in the early phase of training and later switch them off. Another direction could be to add an entropy term to the loss function so that the policy's randomness does not decay too fast.

\paragraph{Nonzero stochasticity} 
We have so far discussed only deterministic LQG environments ($\sigma=0$). As most of industrial use cases of RL assume the presence of stochasticity in environments, it is important to benchmark the proposed methods at finite noise. With this in mind we have run \NAME{} and \NAMEV{} in two-dimensional \emph{stochastic} LQG with $\sigma=1.0$. At each iteration during training, we monitored the performance of the trained policy by running them for 10 test episodes. After training, we sampled 1500 latent variables and fed them into the best policy. Each of the resultant 1500 policies was evaluated on 200 test episodes. This process was repeated for 5 independent runs. The reference value for optimal HV was obtained by solving the Riccati equation numerically for a series of weights; the optimal control policy thus obtained for each weight was tested on 2000 episodes, to reduce statistical errors.

\begin{figure}[t]
	\centering
	\includegraphics[width=.6\columnwidth]{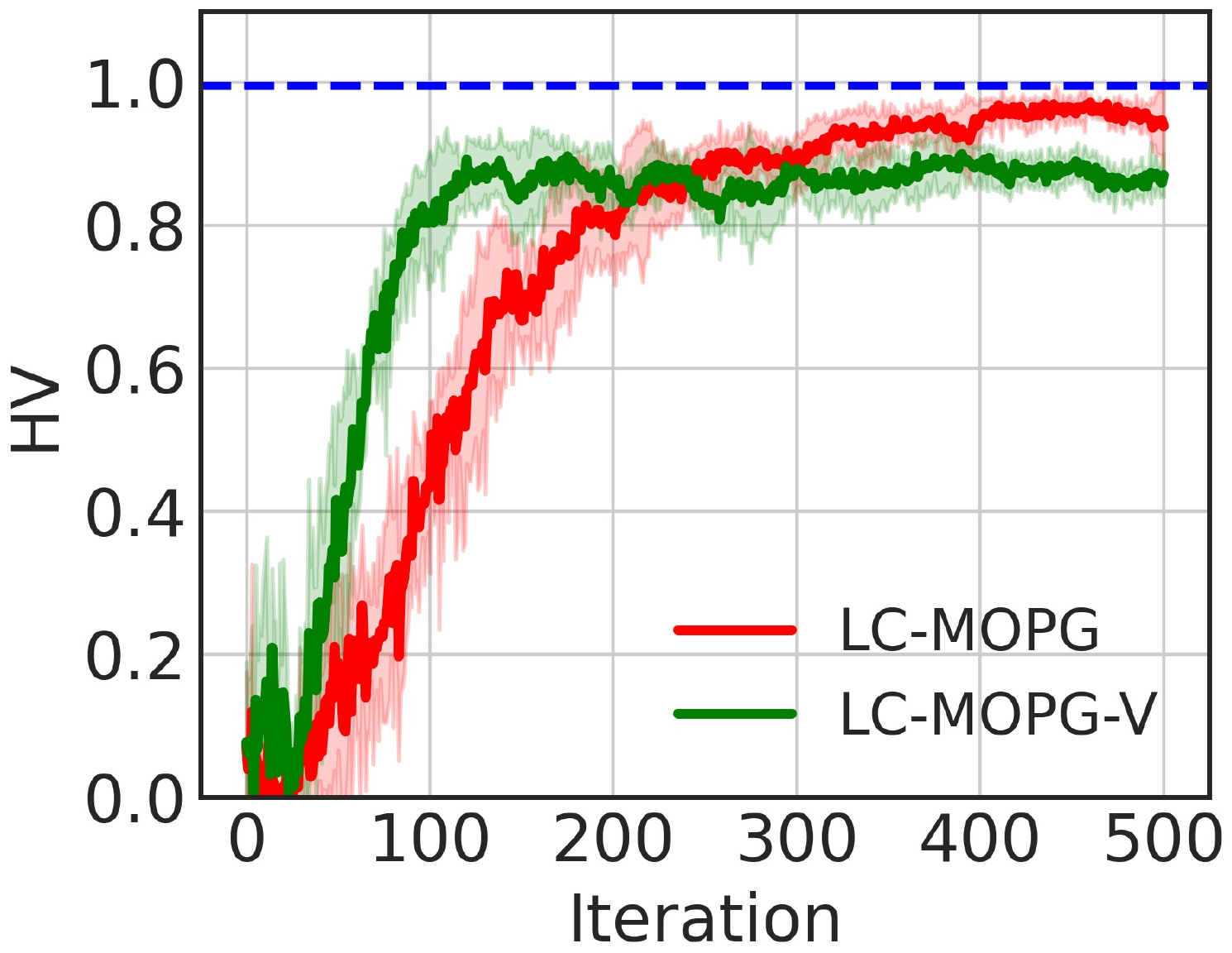}
	\caption{\label{fg_learning_noisy_lqg_2d}Same as Fig.~\ref{fg_learning_lqg_2d} but with nonzero stochasticity $(\sigma=1.0)$.}
\end{figure}

The scores are summarized in Table~\ref{tb:dsmfnprew3}. We note that the optimal HV value has dropped from 1.1646 (Table~\ref{tb:pm45xvr96}) to 0.9967 by 14\% reflecting the effect of disturbance ($\sigma$).%
\begin{table}[tb]
	\centering
	\caption{\label{tb:dsmfnprew3}Comparison of methods in two-dimensional stochastic LQG environment with $\sigma=1.0$.}
	\begin{tabular}{lc}
		\toprule 
		& {\bf HV} $(\uparrow)$
		\\\midrule
		{\bf Optimal value} & 0.9967
		\\
		{\bf \NAME{} $(d_{\rm lat}=2)$} & $0.9616\pm 0.0038$
		\\
		{\bf \NAMEV{} $(d_{\rm lat}=2)$} &  
		$0.9144\pm 0.0245$
		\\\bottomrule
	\end{tabular}
\end{table}
It is satisfactory to see that \NAME{} attained 96.5\% of the optimal HV, implying that a wide and accurate PF was obtained with this method. In contrast, \NAMEV{} attained 91.7\% of the optimal HV. Their learning curves are juxtaposed in Fig.~\ref{fg_learning_noisy_lqg_2d}. 
We observe that the same trend as in Figs.~\ref{fg_learning_lqg_2d} and \ref{fg_learning_lqg_3d} persists here.

\subsection{Results for Minecart}

In this section we discuss numerical results in the Minecart environment. For numerical evaluation of policies we will use the point $(0, 0, -200)$ as a reference point of HV. As shown in \cite{DBLP:conf/atal/ReymondBN22}, the majority of points in the PF lie on the straight line between $(0, 1.5)$ and $(1.5, 0)$ on the plane of two kinds of mined ores due to the fact that the cart's capacity is 1.5. Hence, neglecting the fuel cost, the upper bound on the possible HV would be $(1.5^2/2) \times 200=225$. An elementary calculation using the profiles of 5 mines reveals that only 17 points on the diagonal line are feasible, which further reduces the upper bound on HV to $199.78$. The actual HV of the true PF must be even lower, because the total fuel cost cannot be zero. 

\begin{figure}[t]
	\centering
	\includegraphics[width=.6\columnwidth]{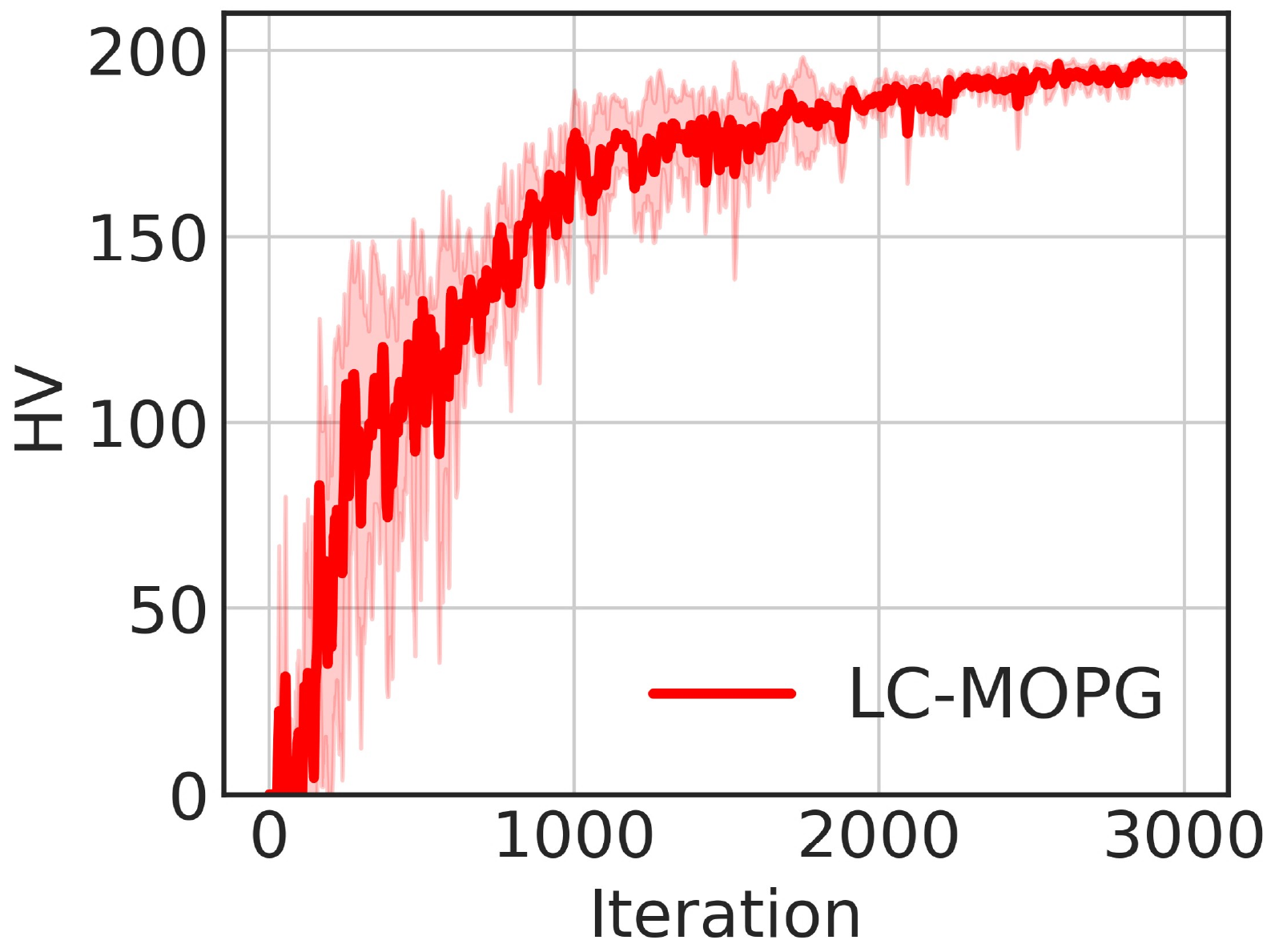}
	\caption{\label{fg_learning_minecart}Training process of \NAME{} in the Minecart environment. To smooth the plot, moving averages over 7 steps were taken.}
\end{figure}

We have trained the agent of \NAME{} using hyperparameter values in Appendix~\ref{sc:mncrt45}. After training, we fed 2000 random latent variables to the policy network and evaluated the HV. The result is summarized in Table~\ref{tb:cspdri3}.  
\begin{table}[tb]
	\centering
	\caption{\label{tb:cspdri3}Comparison of methods in the Minecart environment. Results for baselines are taken from \cite{DBLP:conf/atal/ReymondBN22}. Best value is marked with an asterisk.}
	\begin{tabular}{lc}
		\toprule 
		& {\bf HV} $(\uparrow)$
		\\\midrule
		{\bf RA} \cite{DBLP:conf/ijcnn/ParisiPSBR14} & 
		$123.92 \pm 0.25$
		\\
		{\bf MO-NES} \cite{DBLP:journals/ijon/ParisiPP17} & 
		$123.81 \pm 23.03$
		\\
		{\bf PCN} \cite{DBLP:conf/atal/ReymondBN22} & 
		$197.56 \pm 0.70$
		\\
		{\bf \NAME{} (Ours)} & $198.17 \pm 0.32^*$
		\\\bottomrule
	\end{tabular}
\end{table}
We observe that \NAME{} outperformed all baselines in terms of HV. The attained score of \NAME{} is close to that of PCN \cite{DBLP:conf/atal/ReymondBN22}, but it is noteworthy that the standard deviation of \NAME{} is less than half of that of PCN, indicating salient stability of the proposed approach. The learning curve is shown in Fig.~\ref{fg_learning_minecart}.  
The training with 5 random seeds took approximately 29 hours. The fact that the number of required iterations (3000) is substantially larger than that for simpler environments such as LQG (500$\sim$800) highlights the complexity of control required in the Minecart environment. Due to the limitation on computational resources, we could not fully explore optimal hyperparameter values. Although we speculate that the training of optimally tuned \NAME{} would converge faster, an in-depth examination of this point is beyond the scope of this paper.

\begin{figure}[t]
	\centering
	\includegraphics[width=.6\columnwidth]{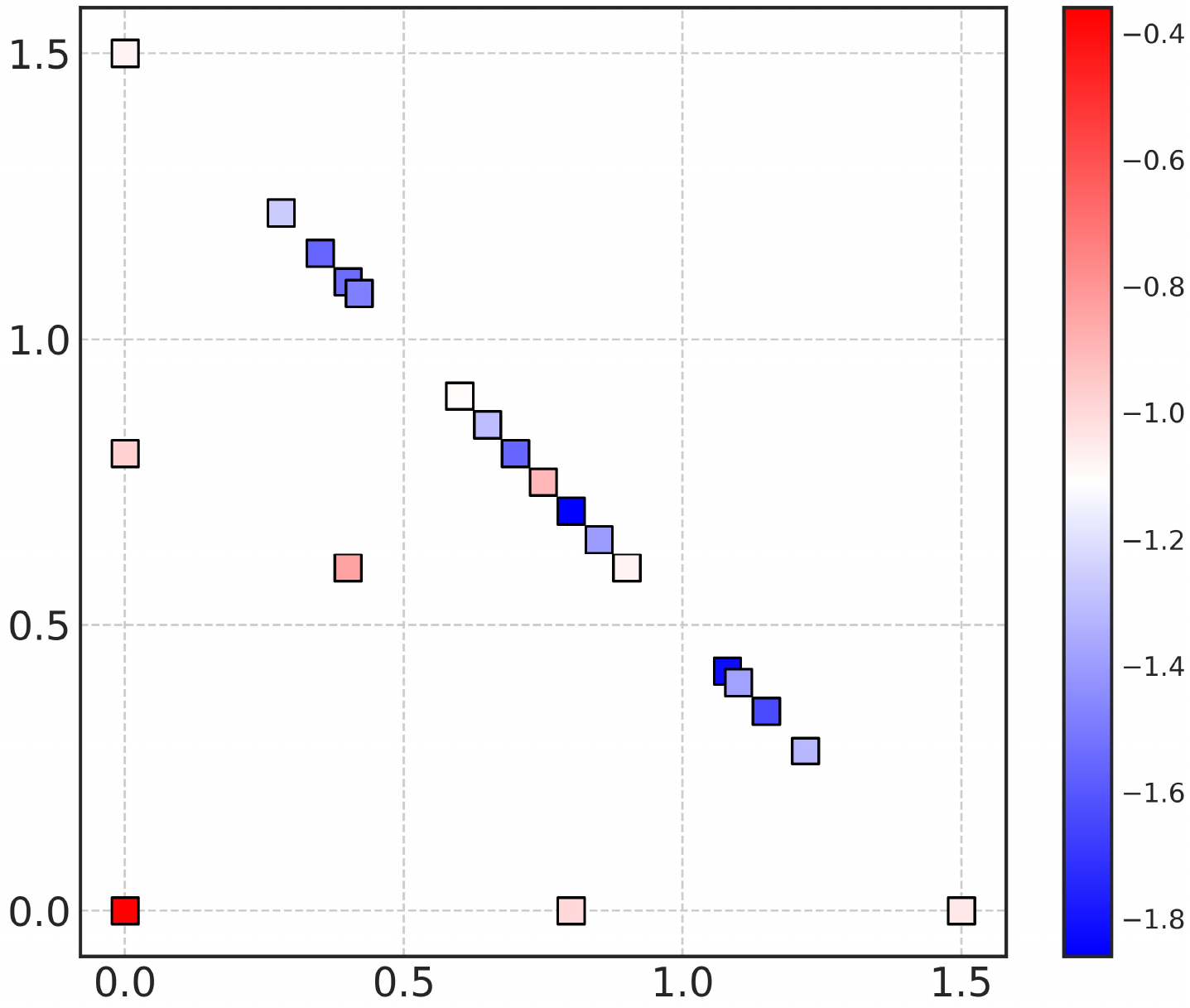}
	\vspace{2mm}
	\caption{\label{fg_minecart_pf}The PF obtained by \NAME{} projected onto the first two dimensions. The color of each point indicates the third component of the return.}
\end{figure}

To gain more insight into the obtained policies, we have plotted the PF of \NAME{} in Fig.~\ref{fg_minecart_pf}. We observe that \NAME{} was able to discover all 17 points on the diagonal line that are associated with trajectories in which the cart mines ores until it gets full. These points are in blue, because their trajectories are generally long and the fuel cost is high (i.e., the third component of return is low).%
\begin{figure*}[t]
	\centering
	\includegraphics[width=.85\textwidth]{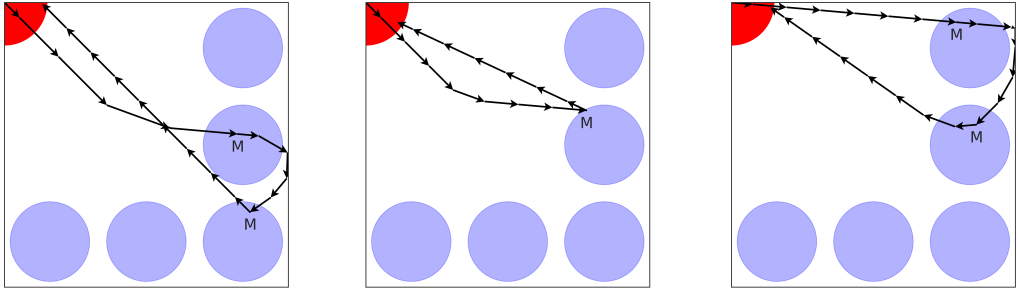}
	\caption{\label{fg_minecart_traj}Example trajectories in the Pareto set. All these trajectories were generated by the same policy network conditioned on different latent values. The value of the return is $(0.65,0.85,-1.30)$ in the left panel, $(0.40,0.60,-0.84)$ in the middle panel, and $(0.28,1.22,-1.38)$ in the right panel, respectively. The symbol `M' indicates that the agent mined ores at that point.}
\end{figure*}
The PF point at the origin $(0,0)$ is associated with an exceptionally short trajectory in which the cart departs home and then returns immediately, without ever reaching mines. The three isolated points below the diagonal represent trajectories in which the cart returns home before it gets full. Although ore1 and ore2 are symmetric in this environment, these three PF points are not symmetric, which implies that a hidden fourth point at $(0.6,0.4)$ has been missed by our agent.

In Fig.~\ref{fg_minecart_traj} we present three example trajectories generated by the trained policy network. It is satisfactory to see that a single policy network can actually produce markedly different behaviors if conditioned on different values of the latent input. 

Finally we comment on the episode length distribution. In the Minecart environment, an episode ends if the cart once departs and later returns home, otherwise the episode will last forever. For numerical efficiency, we set the maximum episode length to 100 during training, and elongated it to 1000 in the evaluation phase. (We count 4 frames as 1 time step.) We collected 10000 random trajectories by running the trained policies from 5 independent runs with random latent input, and found that 8236 trajectories were of length 1000. This means that in more than 80\% of the acquired behaviors, the cart either stays home forever, or departs but never returns home.%
\begin{figure}[t]
	\centering
	\includegraphics[width=.6\columnwidth]{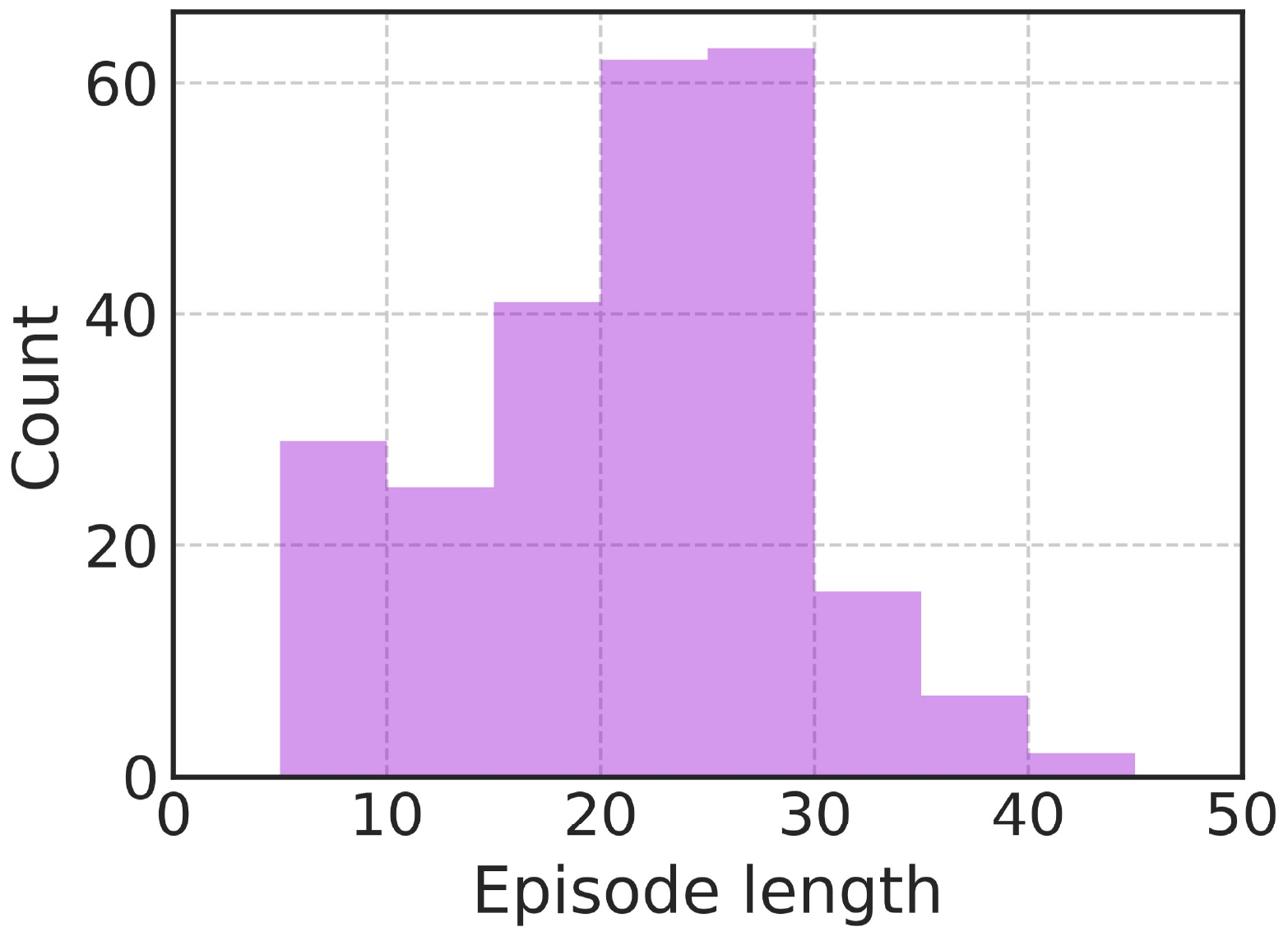}
	\caption{\label{fg_minecart_ep_len}Distribution of the episode lengths of 247 trajectories that correspond to the PF depicted in Fig.~\ref{fg_minecart_pf} in the Minecart environment.}
\end{figure}
Out of the remaining 1764 cases, we found 247 cases to belong to the PF. The episode lengths in these cases obey the distribution in Fig.~\ref{fg_minecart_ep_len}. Most episodes in the PF seem to end with less than 50 steps. This verifies that the maximum episode length limit (100) enforced during training does not interfere with the learning of optimal policies.

\section{Conclusions and outlook\label{sc:452sdfgz}}

Optimal control problems in real life often involve multiple objectives that are hard to maximize simultaneously, such as the speed and safety in car driving. Single-objective RL methods may apply if there is an agreed way to aggregate multiple objectives into a single target by means of, e.g., linear scalarization that uses a user's preference over objectives. However, preference may vary from one occasion to another, and obtaining separate policies through separate training could be quite inefficient. In this work, we introduced \NAME{}, an MORL algorithm that obtains infinitely many policies in a single run of training. It uses a latent-conditioned NN to learn diverse behaviors in a parameter-efficient way. It does not rely on linear scalarization and hence is capable of finding the concave part of the PF. In numerical experiments we confirmed that \NAME{} performed on per with or even better than standard baselines from the MORL literature. While recent work on MORL predominantly focuses on value-based methods (some variants of Q learning) \cite{DBLP:conf/ijcnn/CastellettiPR12,Castelletti2013WRR,DBLP:journals/corr/MossalamARW16,DBLP:conf/icml/AbelsRLNS19,Reymond2019,DBLP:conf/nips/YangSN19,DBLP:journals/corr/abs-2208-07914}, our work demonstrates that policy-based methods still stand as a competitive alternative. 

In future research, we wish to employ the proposed methods in solving complex real-life problems in industry. The challenge for us is then to improve sample complexity. The proposed methods (\NAME{} and \NAMEV{}) collect a large number of rollout trajectories on-the-fly, and make use of them only once to update the policy parameters. It would be valuable if we could introduce some kind of off-policiness so that stored rollout trajectories may be utilized more than once. Another challenge is to evaluate the proposed methods on problems with high-dimensional continuous states such as images, and problems with partially observable states (i.e., POMDP). How to utilize specialized architectures such as convolutional NN and recurrent NN as a component of the policy network for \NAME{} is an interesting question that calls for further investigation.

\begin{appendices}
\section{Hyperparameter Settings}
\subsection{Hyperparameters for DST\label{sc:4332}}
\begin{center}
\begin{tabular}{cc}
	\toprule
	Hyperparameter & Value
	\\\midrule
	$d_{\rm lat}$ & 3
	\\
	$N_{\rm lat}$ (train and test) & 400
	\\
	width of policy's MLP & 36
	\\
	depth of policy's MLP & 3
	\\
	max episode length & 50
	\\
	$k$ & 10
	\\
	$\beta$ & 4.0
	\\
	$\gamma$ (train and test) & $\left\{\begin{array}{c}
	    \!\! 0.99\; \text{(convex case)}
	    \\
	    \!\! 1.0\; \text{(original case)}
	\end{array}\right.$
	\\
	normalization of returns & Max-min
	\\
	number of iterations & 30
	\\\bottomrule
\end{tabular}
\end{center}

\subsection{Hyperparameters for FTN $(d=5)$ \label{sc:fktrdfg89}}
\begin{center}
\begin{tabular}{cc}
	\toprule 
	Hyperparameter & Value
	\\\midrule
	$d_{\rm lat}$ & 5 
	\\
	$N_{\rm lat}$ (train and test) & 300 
	\\
	width of policy's MLP & 100 
	\\
	depth of policy's MLP & 3
	\\
	max episode length & ---
	\\
	$k$ & 3
	\\
	$\beta$ & 5.0
	\\
	$\gamma$ (train and test) & 0.99
	\\
	normalization of returns & Max-min
	\\
	number of iterations & 20
	\\
	$(e1, e2)$ & $(10, 20)$
	\\\bottomrule 
\end{tabular}
\end{center}

\subsection{Hyperparameters for FTN $(d=6)$ \label{sc:fkerw9}}
\begin{center}
\begin{tabular}{cc}
	\toprule 
	Hyperparameter & Value
	\\\midrule
	$d_{\rm lat}$ & 7 
	\\
	$N_{\rm lat}$ (train) & 400 
	\\
	$N_{\rm lat}$ (test) & 1500
	\\
	width of policy's MLP & 140 
	\\
	depth of policy's MLP & 3
	\\
	max episode length & ---
	\\
	$k$ & 10
	\\
	$\beta$ & 10.0
	\\
	$\gamma$ (train and test) & 0.99
	\\
	normalization of returns & Max-min
	\\
	number of iterations & 20
	\\
	$(e1, e2)$ & $(10, 10)$
	\\\bottomrule 
\end{tabular}
\end{center}

\subsection{Hyperparameters for FTN $(d=7)$ \label{sc:8ewr39}}
\begin{center}
\begin{tabular}{cc}
	\toprule 
	Hyperparameter & Value
	\\\midrule
	$d_{\rm lat}$ & 7 
	\\
	$N_{\rm lat}$ (train) & 400 
	\\
	$N_{\rm lat}$ (test) & 1500 
	\\
	width of policy's MLP & 210 
	\\
	depth of policy's MLP & 3
	\\
	max episode length & ---
	\\
	$k$ & 10
	\\
	$\beta$ & 10
	\\
	$\gamma$ (train and test) & 0.99
	\\
	normalization of returns & Max-min
	\\
	number of iterations & 20
	\\
	$(e1, e2)$ & $(10, 10)$
	\\\bottomrule 
\end{tabular}
\end{center}

\subsection{Hyperparameters for LQG (2D) \label{sc:lqg2}}
\begin{center}
\begin{tabular}{cc}
	\toprule 
	Hyperparameter & Value
	\\\midrule
	$d_{\rm lat}$ & 1,~2,~3
	\\
	$N_{\rm lat}$ (train) & 200 
	\\
	$N_{\rm lat}$ (test) & 1500 
	\\
	width of policy's MLP & 24 
	\\
	depth of policy's MLP & 3
	\\
	max episode length & 30 
	\\
	$k$ & 3 
	\\
	$\beta$ & 10.0 
	\\
	$\gamma$ (train and test) & 0.9
	\\
	normalization of returns & Robust  
	\\
	number of iterations & 500  
	\\\bottomrule 
\end{tabular}
\vspace{\baselineskip}
\\
\begin{tabular}{cc}
	\toprule
	Hyperparameter for $Q$ and $V$ & Value
	\\\midrule 
	epochs per iteration & 1
	\\
	batch size & 64
	\\
	hidden layer width & 24
	\\
	hidden layer depth & 3
	\\\bottomrule
\end{tabular}
\end{center}

\subsection{Hyperparameters for LQG (3D) \label{sc:lqg3}}
\begin{center}
\begin{tabular}{cc}
	\toprule 
	Hyperparameter & Value
	\\\midrule
	$d_{\rm lat}$ & 1,~2,~3
	\\
	$N_{\rm lat}$ (train) & 300 
	\\
	$N_{\rm lat}$ (test) & 1500 
	\\
	width of policy's MLP & 30 
	\\
	depth of policy's MLP & 3
	\\
	max episode length & 30 
	\\
	$k$ & 3 
	\\
	$\beta$ & 10.0
	\\
	$\gamma$ (train and test) & 0.9
	\\
	normalization of returns & Robust  
	\\
	number of iterations & 800  
	\\\bottomrule 
\end{tabular}
\vspace{\baselineskip}
\\
\begin{tabular}{cc}
	\toprule
	Hyperparameter for $Q$ and $V$ & Value
	\\\midrule 
	epochs per iteration & 1
	\\
	batch size & 100
	\\
	hidden layer width & 30 
	\\
	hidden layer depth & 3
	\\\bottomrule
\end{tabular}
\end{center}

\subsection{Hyperparameters for Minecart \label{sc:mncrt45}}
\begin{center}
\begin{tabular}{cc}
	\toprule 
	Hyperparameter & Value
	\\\midrule
	$d_{\rm lat}$ & 3 
	\\
	$N_{\rm lat}$ (train) & 400 
	\\
	$N_{\rm lat}$ (test) & 2000 
	\\
	width of policy's MLP & 36 
	\\
	depth of policy's MLP & 3
	\\
	max episode length & $\left\{\begin{array}{c}
	    \!\! 100\; \text{(train)}
	    \\
	    \!\! 1000\; \text{(test)}
	\end{array}\right.$
	\\
	$k$ & 3
	\\
	$\beta$ & 6.0
	\\
	$\gamma$ (train and test) & 1.0
	\\
	normalization of returns & Max-min 
	\\
	number of iterations & 3000
	\\
	$(e1, e2, e3, e4, e5, e6)$ & all 10
	\\\bottomrule 
\end{tabular}
\end{center}
\end{appendices}

\section*{Declarations}
\subsection*{Conflict of interest} 
The authors declare that they have no conflict of interest.
\subsection*{Ethical approval} 
This article does not contain any studies with animals performed by any of the authors.

\bibliography{draft_v3.bbl}
\end{document}